\documentclass[lettersize,journal]{IEEEtran}
\usepackage{amsmath,amsfonts,amssymb}
\usepackage{algorithmic}
\usepackage{array}
\usepackage{graphicx}
\usepackage{textcomp}
\usepackage{stfloats}
\usepackage{booktabs}
\usepackage{multirow}
\usepackage{tabularx}
\usepackage{float}
\usepackage{mdwmath}
\usepackage{mdwtab}
\usepackage{eqparbox}
\usepackage{url}
\usepackage{xcolor}
\usepackage{soul}

\usepackage[caption=false,font=normalsize,labelfont=sf,textfont=sf]{subfig}

\definecolor{myColor}{rgb}{0.8039,0,0}

\usepackage{hyperref}
\def\BibTeX{{\rm B\kern-.05em{\sc i\kern-.025em b}\kern-.08em
    T\kern-.1667em\lower.7ex\hbox{E}\kern-.125emX}}
\usepackage{balance}
\begin{document}
\title{PolyBuild: An End-to-End Method for Polygonal Building Contour Extraction from High-Resolution Remote Sensing Images}
\author{Yaoteng Zhang, Julin Zhang, Guangshuai Wang, Jiwei Deng, Hui Sheng, Yasir Muhammad, 

Shiqing Wei\,~\IEEEmembership{Member,~IEEE}
\thanks{This work was supported in part by the Tianjin Key Laboratory of Rail Transit Navigation Positioning and Spatio-temporal Big Data Technology (Grant No. TKL2024A11);in part by the  National Natural Science Foundation of China (Grant No. 42401428); in part by the Postdoctoral Fellowship Program of CPSF (Grant No. GZB20240849); in part by the Natural Science Foundation of Shandong Provincial (Grant No. ZR2024QD133); in part by Qingdao Postdoctoral Funding Program (Grant No. ODBSH20240201022). (Corresponding authors: Shiqing Wei.)

Yaoteng Zhang, Hui Sheng, Yasir Muhammad, Shiqing Wei are with the College of Oceanography and Space Informatics, China University of Petroleum (East China), Qingdao 266580, China. (e-mail: zhang\_yt@s.upc.edu.cn; sheng@upc.edu.cn; LB2116001@s.upc.edu.cn; wei\_sq@upc.edu.cn;)

Julin Zhang is with the South Surveying\&Mapping Instrument Co.,Ltd. (e-mail: julin.zhang@southgis.com)

Guangshuai Wang, Jiwei Deng are with the China Railway Design Corporation, Tianjin, China. (e-mail: gswang0806@126.com; dengjiwei@crdc.com)}}

% \markboth{Journal of \LaTeX\ Class Files,~Vol.~18, No.~9, September~2020}%
% {How to Use the IEEEtran \LaTeX \ Templates}

\maketitle

\begin{abstract}
% Extracting building polygon contours from high-resolution remote sensing images is crucial for mapping applications. However, varying imaging conditions and complex building structures make automatic contour extraction challenging. We propose an end-to-end method named PolyBuild which can directly extract building vector polygons without any post-processing operations. PolyBuild includes an Initial Contour Generation Module (ICGM), which utilizes concatenated sub-region center features to generate initial contours for each building instance, and a Contour Optimization Module (COM), which iteratively refines building contours by integrating CNN and Transformer architectures to achieve more accurate polygonal shapes. The ICGM preforms simultaneous object detection and contour extraction. By generating bounding boxes to obtain sub-region center points, ICGM concatenates the features of four sub-region centers to generate initial building contours. The COM combines CNN features and contour positional information in a Transformer decoder, facilitating global and local information extraction, automatically capturing dependencies between different vertices and iteratively updating the contours for precise polygon extraction. Extensive experiments on the WHU aerial building dataset and WHU-Mix dataset demonstrate that PolyBuild achieves higher accuracy compared to other mask-based and contour-based methods, establishing a new state-of-the-art (SOTA).
Extracting building polygon contours from high-resolution remote sensing images is a fundamental task for various mapping applications. However, the presence of varying imaging conditions and complex building structures, makes automatic contour extraction extremely challenging. Mainstream approaches for building extraction often rely on pixel-level segmentation followed by multiple post-processing steps to produce building contour, which can be computationally intensive and prone to errors. In this paper, we propose an end-to-end method named PolyBuild, which can directly extract building vector polygons from high-resolution remote sensing images without the need for any post-processing operations. The proposed method leverages two primary modules: an Initial Contour Generation Module (ICGM) and a Contour Optimization Module (COM). The ICGM is designed to generate an initial building contour by utilizing concatenated sub-region center features for each building instance. It performs simultaneous object detection and initial contour extraction by generating bounding boxes and using the center features of four sub-regions to represent each building. The Contour Optimization Module (COM) further refines the generated building contours by iteratively integrating Convolutional Neural Network (CNN) features and contour positional information in a Transformer-based decoder. The hybrid CNN-Transformer architecture effectively captures both local and global spatial relationships within the building contour, ensuring high-quality boundary delineation. Extensive experiments are conducted on three building datasets to evaluate the performance of PolyBuild. The results demonstrate that PolyBuild significantly outperforms state-of-the-art methods, including mask-based and contour-based approaches.
\end{abstract}

\begin{IEEEkeywords}
Building Contour Extraction, Remote Sensing Images, CNN-Transformer, Instance Segmentation
\end{IEEEkeywords}

\section{Introduction}
\IEEEPARstart{E}{xtracting} vector representations of building polygons from aerial and satellite images is increasingly important in many applications, such as cartography, urban modeling and reconstruction, and map generation \cite{zorzi2022polyworld}. However, manually annotating buildings in remote sensing imagery is a time-consuming and labor-intensive task.

To improve mapping efficiency, previous methods based on hand-crafted features aim to find standard representations of building appearances, such as color, shape, shadow, and texture \cite{xu2021gated,li2014extracting,guo2020scene}. Among these, corner detection operators and edge detection operators \cite{hart2000pattern,shrivastava2015automatic,canny1986computational} are widely used. However, various detection operators are limited in their applicability due to obstructions from trees and shadows. Methods \cite{yunfeng2018extraction,huang2011multidirectional} based on auxiliary information combined with multi-source remote sensing data, such as digital surface model (DSM) \cite{yu2021automatic}, enrich the semantic information of buildings, but these methods are very time-consuming and expensive. Methods relying on hand-crafted features and expert experience often target specific types of buildings, demonstrating limited generalization capabilities in large-scale real-world applications. 

Unlike traditional methods that heavily rely on manual feature selection, approaches based on deep learning can automatically learn features from data, thus demonstrating improved robustness \cite{guo2023decoupling,yasir2024shipgeonet}. Mask-based methods are commonly employed for building extraction tasks, which involve classifying pixels into two categories: buildings (the target class) and non-buildings \cite{10824925}. However, these methods present several limitations. Firstly, mask-based approaches do not provide direct access to the vector contour information of buildings, which is often required in practical applications. Secondly, the contours of buildings extracted using segmentation methods are frequently rough and jagged. Lastly, when faced with obstacles such as trees and shadows, segmentation-based methods struggle to fully extract building counts \cite{wei2019toward,zhou2022bomsc,zhu2020map,hu2023boundary,he2017mask,bolya2019yolact}.

% Semantic segmentation methods \cite{wei2019toward,zhou2022bomsc,zhu2020map,hu2023boundary} are widely used in building extraction tasks, classifying images pixel by pixel into buildings (target class) and non-buildings. However, semantic segmentation methods are unable to extract individual building information.

% Instance segmentation methods, such as Mask R-CNN \cite{he2017mask} and Yolact \cite{bolya2019yolact}, achieve instance-level information extraction by introducing bounding boxes. However, the essence of these methods remains pixel-level classification, resulting in poor contour quality and an inability to directly obtain vectorized contours of the targets. 

\begin{figure}[]
    \centering
    \includegraphics[width=9cm]{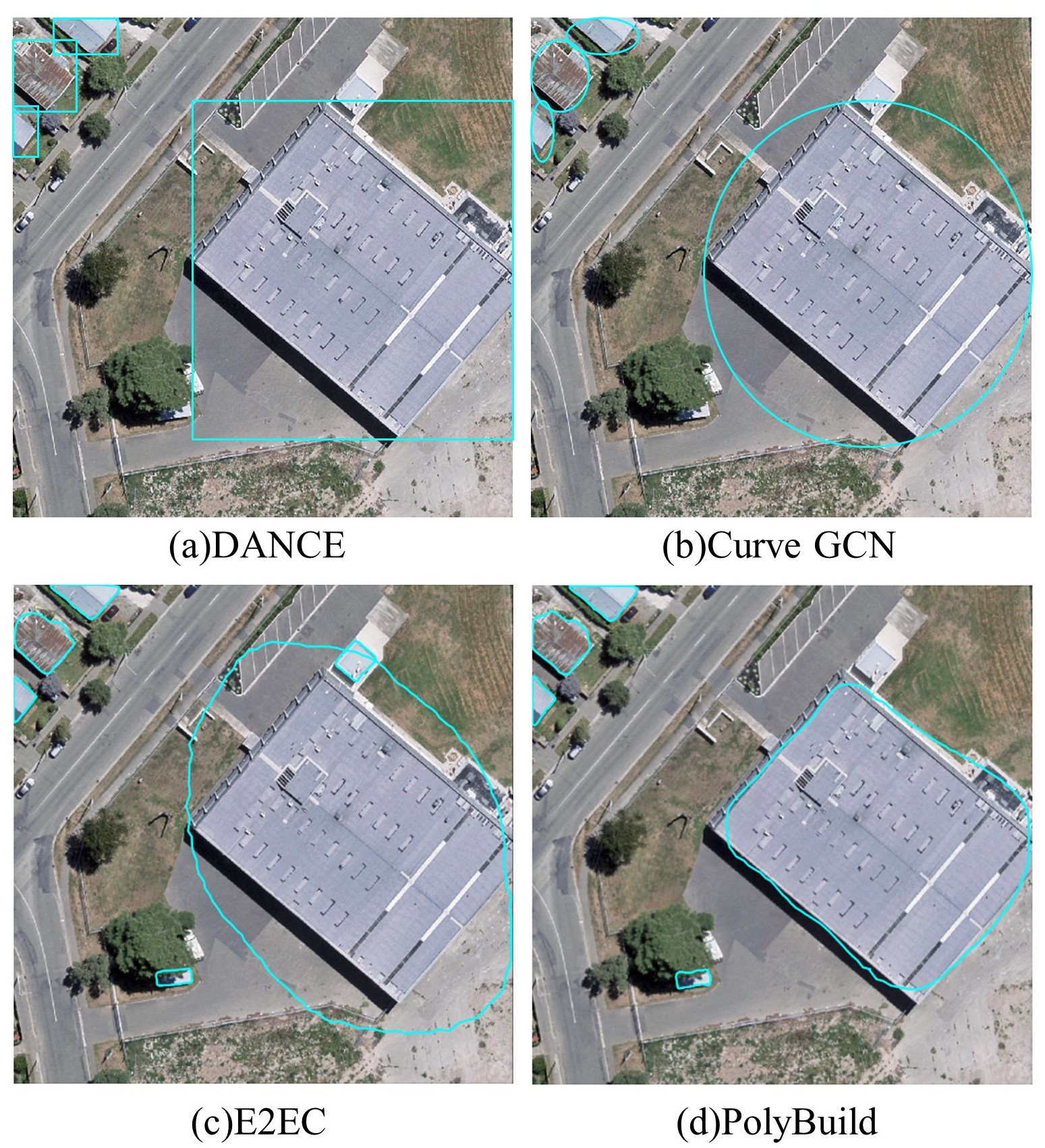}
    \vspace{-2em} % 根据需要调整这个值
    \caption{Different methods generate initial contours as follows: (a) DANCE \cite{liu2021dance} uses bounding boxes, (b) Curve GCN \cite{ling2019fast} utilizes ellipses, (c) E2EC \cite{zhang2022e2ec} adapts initial contours using center point features, (d) PolyBuild generates initial contours based on coupled features of sub-region center points.}
    \label{fig:1}
\end{figure}

Compared mask-based methods, the contour-based methods \cite{xie2020polarmask,peng2020deep,liu2021dance,zhang2022e2ec,wei2023buildmapper} consider instance extraction as a contour regression task, generating target contours by regressing a series of contour vertices. Compared to mask-based methods, the contour-based methods theoretically achieves higher efficiency and can directly obtain vectorized building contours without the need for post-processing operations. However, existing contour-based methods still have several notable drawbacks. Firstly, As shown in Fig. \ref{fig:1}(a)(b), DANCE \cite{liu2021dance} and Curve GCN \cite{ling2019fast} use bounding boxes and ellipses as initial contours, which are far from the true target contours. Although E2EC \cite{zhang2022e2ec}, shown in Fig. \ref{fig:1}(c), employs building center features to generate initial contours, thereby eliminating the need for manual initialization, due to the use of limited features to regress the building initial contour, as the regression distance increases , the predicted building contours may contain erroneous information. To address this issue, we introduce an Initial Contour Generation Module (ICGM), which regresses the initial building contour from the coupled sub-regions center features. Compared to using  single center point features, the coupled sub-region center features provide richer building information. Moreover, the sub-region center points are closer to the building contour in terms of relative coordinate position, significantly reducing the difficulty of generating building contours.

Secondly, Multi-stage methods \cite{peng2020deep,liu2021dance,ling2019fast} use contour features to iteratively adjust the rough initial contour. However, the convolution-based method can only extracts local feature information between fixed vertices and cannot automatically implement global correlation feature indexing, resulting in inaccurate regression offsets. To address this problem, we design a Contour Optimization Module (COM) that integrates CNN-Transformer architectures to extract both local and global information and iteratively optimize the initial contour. The COM obtains local features of contour vertices through CNN components, and then connects local features  with vertex position mapping features as input to the Transformer decoder, thus realizing the association between vertex position information and features. The Transformer decoder uses the attention module to query global vertices and establish information association between vertex features at different positions, thereby correcting contour errors and optimizing contour polygons more accurately.

In summary, our contributions are as follows:

•	We propose an end-to-end building contour extraction framework, PolyBuild. This framework enables the parallel implementation of building detection and contour extraction. Experiments on three building datasets demonstrate state-of-the-art performance of PolyBuild on instance-level contour extraction.

•	We design a powerful and innovative Initial Contour Generation Module (ICGM) for generating initial contours by utilizing coupled sub-region center point features. ICGM produces more stable and precise initial contours based on the characteristics of the center points from four sub-regions within the bounding box, significantly reducing the complexity of subsequent contour optimization.

•	We innovatively integrate the CNN-Transformer architecture to propose a Contour Optimization Module (COM), which can automatically query and index associated vertices, better understand vertex positions, and achieve accurate contour prediction.

\section{Related work}
{\bf{Mask-based methods.}} Semantic segmentation methods, particularly classic CNN-based approaches like U-Net \cite{ronneberger2015u} and HRNet \cite{sun2019deep}, are extensively utilized for building extraction tasks. These methods enhance segmentation accuracy by extracting multi-scale features through the gradual fusion of feature maps across layers. Some studies \cite{zhu2020map,hu2023boundary,jung2021boundary,xu2023bctnet} have refined network modules to improve sensitivity to building edges and enhance edge information for better recognition accuracy. Transformer architectures leverage attention mechanisms to capture global context, with Vision Transformer (ViT) \cite{dosovitskiy2020image} and SegFormer \cite{xie2021segformer} showcasing the potential of Transformers in computer vision tasks. However, semantic segmentation methods cannot provide information on individual buildings, limiting their applicability in downstream industries. Mask R-CNN \cite{he2017mask} and PANet \cite{liu2018path} integrate segmentation tasks within an object detection framework, first generating bounding boxes through object detection and then performing pixel-wise segmentation within these boxes. While this approach can yield performance improvements, it often results in slower processing speeds. YOLACT \cite{bolya2019yolact}, CondInst \cite{tian2020conditional}, and BlendMask \cite{chen2020blendmask} incorporate a mask branch into single-stage object detection models, striking a balance between accuracy and speed. To obtain vectorized contours, raster masks typically require post-processing through vectorization \cite{ji2018fully}. Nevertheless, the building footprint extracted by mask-based methods often exhibit rough contours and significant jaggedness, and these methods may yield incorrect predictions when obstructed by trees and shadows.

{\bf{Contour-based methods.}} Contour-based methods conceptualize instance segmentation as a contour point regression problem. Unlike mask-based methods, contour-based approaches can directly derive the vectorized contour of the target object. Some studies \cite{huang2021sequentially, zhao2021building, liu2022building} employ Recurrent Neural Networks (RNNs) to sequentially predict vertices for generating building contours; however, these methods are often susceptible to error accumulation. Methods such as Hisup \cite{xu2023hisup}, SAMPolyBuild \cite{wang2024sampolybuild}, and PolyCity \cite{li2023joint} aid in extracting building vertices by predicting segmentation maps and subsequently reconstruct the adjacency relationships between these vertices to restore the building's vector contour. Inspired by Polyworld \cite{zorzi2022polyworld}, Line2Poly\cite{wei2024lines} reconstructs building contours by restoring the topological relationships among building feature lines. PolarMask \cite{xie2020polarmask} directly regresses the coordinates of instance contour vertices based on center point features, but this approach struggles with complex concave objects. Although Deepsnake \cite{peng2020deep}, DANCE \cite{liu2021dance}, and Curve GCN \cite{ling2019fast} utilize contour features for extraction, the predefined initial contour shapes can limit model performance. E2EC \cite{zhang2022e2ec} and Buildmapper \cite{wei2023buildmapper} generate initial contours from center point features, significantly enhancing the applicability of the network. However, as the regression distance increases, these methods often struggle to accurately restore the initial contour based on limited center features, which may introduce erroneous information. The accuracy of the initial contour is crucial for the subsequent optimization of the contour. Therefore, our motivation is to generate higher-quality initial contours to achieve more accurate contour optimization.

% The contour-based methods view instance segmentation as a contour points regression problem. Compared to mask-based methods, contour-based methods can directly obtain vectorized contours of the target objects. Studies \cite{huang2021sequentially,zhao2021building,liu2022building} generate building contours by sequentially predicting vertices using RNNs, but these approaches are often prone errors accumulation. PolarMask \cite{xie2020polarmask} directly regresses the instance contour vertex coordinates based on center point features. However, this method performs poorly when dealing with complex concave objects. Although Deepsnake \cite{peng2020deep}, DANCE \cite{liu2021dance}, and Curve GCN \cite{ling2019fast} utilize contour features to achieve contour extraction, the artificially set initial contour shape limits the performance of the model. E2EC \cite{zhang2022e2ec} generates the initial contour by center point features which greatly increases the applicability of the network. However, as the regression distance increases, E2EC struggles to effectively restore the initial contour based on limited center features, and is prone to introducing erroneous information.

\section{Methodology}
 \begin{figure*}[]
    \centering
    \includegraphics[width=1\linewidth]{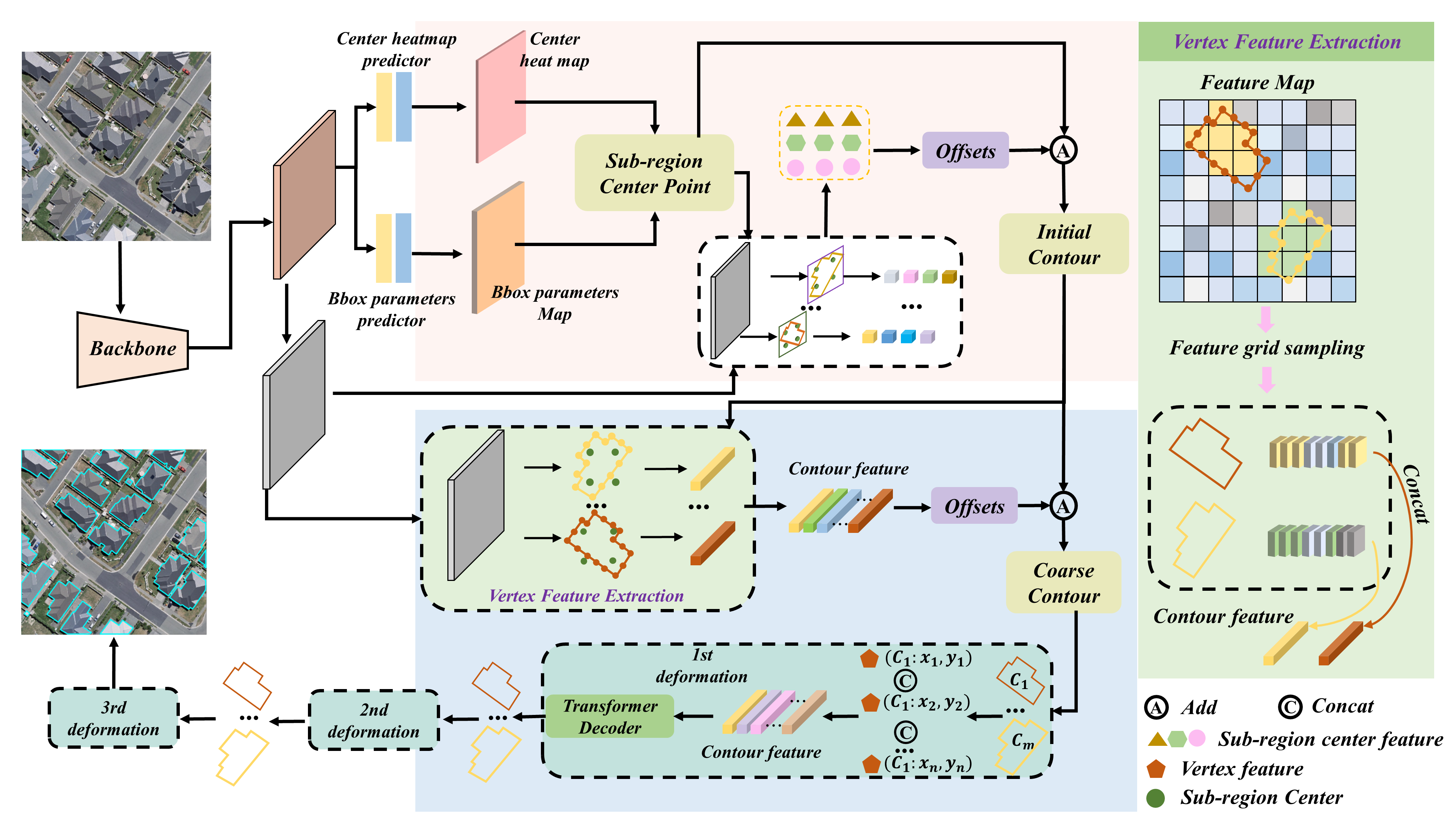}
    \caption{Overall architecture of the PolyBuild. PolyBuild generates the initial contour through Initial Contour Generation Moduel (ICGM) and then inputs the initial contour into Contour Optimization Module (COM) for iterative optimization to output contour polygons.}
    \label{fig:overall}
\end{figure*}
\subsection{Overview}
The proposed framework, PolyBuild, seamlessly integrates building detection and building contour polygon extraction into an end-to-end framework. As shown in Fig. \ref{fig:overall}, PolyBuild comprises two primary modules: the Initial Contour Generation Module (ICGM) and the Contour Optimization Module (COM). Given an image, the ICGM autonomously generates initial building contours. These preliminary contours are then subsequently processed by the COM, which employs a CNN-Transformer architecture for iterative refinement across three iterations, ultimately producing accurate building contour polygons.

\subsection{Initial Contour Generation Module}

\begin{figure*}[]
    \centering
    \includegraphics[width=18cm]{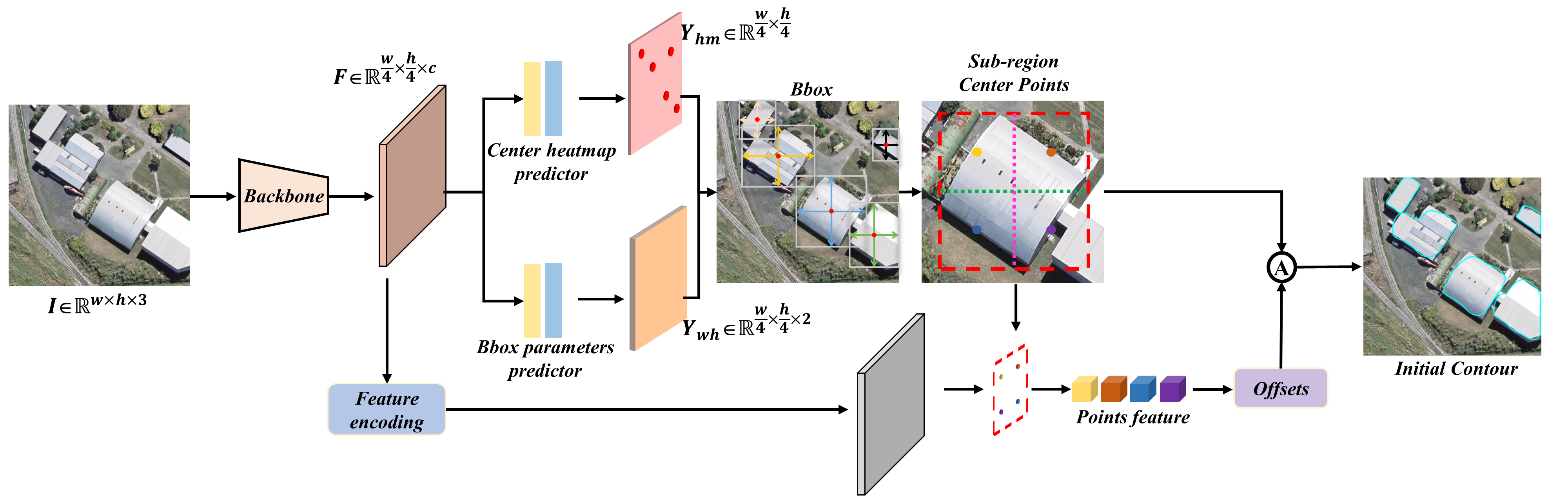}
    \caption{The architecture of the Initial Contour Generation Module (ICGM). First, the ICGM predicts a bbox for each building instance and divides the bbox into four sub-regions. Then, the sub-regions center feature are concatenated to predict a set of vertex offsets. Finally, the offsets are added to the center points of the sub-regions to generate the initial contour.}
    \label{fig:icgm}
\end{figure*}

The Initial Contour Generation Module (ICGM), illustrated in Fig. \ref{fig:icgm}, begins by generating a bounding box (bbox) for each building instance and dividing it into four sub-regions along the center points. Subsequently, the initial building contour vertices are regressed from the feature of sub-region centers. The ICGM comprises the following components: the DLA-34 backbone network, the building center prediction head, the bbox parameters prediction head, and the offsets prediction head.
 
\subsubsection{Definition of Sub-region Center Points}

Regressing building contours from limited center point features is challenging, especially as the regression offsets increase, exacerbating the uncertainty of the initial contour. Therefore, we propose an Initial Contour Generation Module (ICGM) coupled with sub-region center features. Firstly, the center prediction head and the bbox parameters prediction head generate the building location information \(Y_{hm}\in[0,1]^{\frac{w}{4} \times \frac{h}{4}} \) and the bbox parameters information (i.e. width and height) \(Y_{wh}\in\mathbb{R}^{\frac{w}{4} \times \frac{h}{4} \times 2}\). Subsequently, the positions of the top \(K \) peaks are extracted from \( Y_{hm} \) as \(b_{ct}\in \mathbb{R}^{K \times 2} \), and the corresponding bbox width and height of the buildings is extracted from \( Y_{wh} \) as \( wh^{pre} \in \mathbb{R}^{K \times 2} \). The building bbox are then generated by combining \( b_{ct} \) and \( wh^{pre} \). Finally, we divide the bbox into four sub-regions (top-left, top-right, bottom-left, and bottom-right) along the center points, and simultaneously obtain the centers of the four sub-regions. Assuming the building center position is \( (x_i, y_i) \) and the corresponding building width and height information is \( (w_i, h_i) \), the four corresponding sub-region center points \( (x_i^{lt}, y_i^{lt}), (x_i^{lb}, y_i^{lb}), (x_i^{rt}, y_i^{rt}), (x_i^{rb}, y_i^{rb}) \) can be obtained by calculation. Take the center of the left-top sub-region as an example:

\begin{equation}
\label{deqn_ex1a}
\begin{cases}
x_i^{lt} = x_i - \frac{1}{4} \times w_i \\
y_i^{lt} = y_i - \frac{1}{4} \times h_i
\end{cases}
\end{equation}
\((x_i^{lt}, y_i^{lt}), (x_i^{lb}, y_i^{lb}), (x_i^{rt}, y_i^{rt}), (x_i^{rb}, y_i^{rb})\) respectively represent the coordinate information of the left-top, left-bottom, right-top, and right-bottom sub-region centers, where \(i\) represents different buildings.

During training, we set \(K\) to 200. It is worth noting that there are few building center pixels compared to other pixels in the image. To mitigate the training difficulties caused by the imbalance between positive and negative samples, Focal loss \cite{lin2017focal} is used for center point prediction head training, and the parameter settings are as CenterNet \cite{zhou2019objects}.

% \minew{Focal Loss addresses the class imbalance problem by introducing a weighting factor $(1 - \hat{Y})^\alpha$, where $\hat{Y}$ is the probability of the predicted center heatmap, and $\alpha$ is an adjustment factor. This weighting factor modulates the influence of the loss function. By reducing the influence of easy-to-classify samples, Focal Loss enables the model to focus more on challenging samples, thereby improving its ability to detect minority class samples, especially in scenarios with significant class imbalance.
% }
Focal Loss addresses the class imbalance problem by introducing a weighting factor $(1 - \hat{Y})^\alpha$, where $\hat{Y}$ is the probability of the predicted center heatmap, and $\alpha$ is an adjustment factor. This weighting factor modulates the influence of the loss function. By reducing the influence of easy-to-classify samples, Focal Loss enables the model to focus more on challenging samples, thereby improving its ability to detect minority class samples, especially in scenarios with significant class imbalance.
 %This approach effectively improves the model's ability to detect minority class samples, especially in situations where the class distribution is highly imbalanced.

% \begin{small} 
\begin{equation}
\label{deqn_ex1a}
L_{ht} = -\frac{1}{M} \sum \left\{
\begin{array}{ll}
(1 - \hat{Y})^{\alpha} \log(\hat{Y}), {if \ } Y = 1 \\
(1 - \hat{Y})^{\beta} (\hat{Y})^{\alpha} \log(1 - \hat{Y}), {otherwise}
\end{array}
\right.
\end{equation}
% \end{small}
\(Y\) represents the ground truth in the center heatmap, \( \alpha \) and \( \beta \) are constants set to 2 and 4 respectively, and \( M \) is the number of center points.

Smooth L1 loss is used to supervise the building bbox parameters prediction head. This loss can reduce the sensitivity to contours while maintaining the differentiability of each bbox and ensuring the smoothness of gradient propagation. The loss calculation of the bbox parameters can be defined as:
\begin{equation}
\label{deqn_ex1a}
Smooth \ L1(x) = \left\{
\begin{array}{ll}
0.5x^2 & {if} \ |x| < 1 \\
|x| - 0.5 & {otherwise}
\end{array}
\right.
\end{equation}
\begin{equation}
\label{deqn_ex1a}
\begin{array}{ll}
L_{wh} = l_1 (wh_i^{pre} - wh_i^{gt})
\end{array}
\end{equation}
where \(l_1\) represents the smooth L1 loss function, \(wh_i^{pre}\) represents predicted bbox parameters amd \( wh_i^{gt}\) represents ground truth bbox parameters.
% where \(l_1\) represents the smooth L1 loss function, \(wh_i^{pre}\) represents predicted bbox parameters amd \( wh_i^{gt}\) represents ground truth bbox parameters.

\subsubsection{Regression of Initial Contours from Sub-region Center Points}
\begin{figure}[]
    \centering
    \includegraphics[width=9cm]{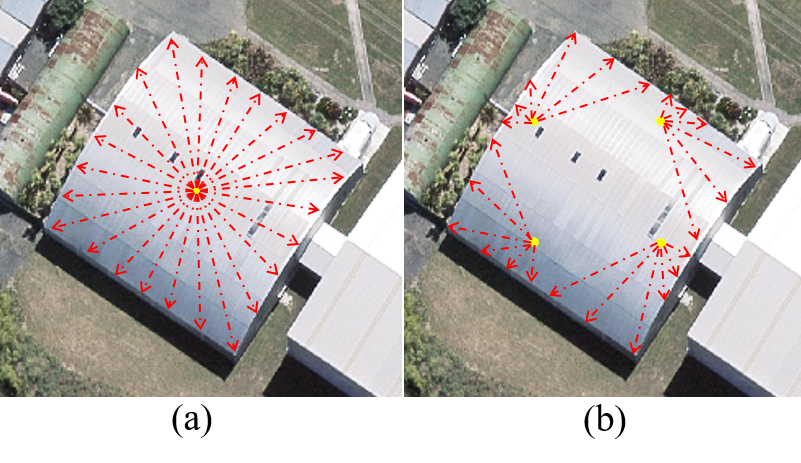}
    \vspace{-2.5em} % 根据需要调整这个值
    \caption{Comparison of building contour vertex offset regression between center point and sub-region center point. (a) Regresses building contour vertices from the center point. (b) Regresses building contour vertices from the sub-region center point.}
    \label{fig:4}
\end{figure}

The building bbox is divided into four sub-regions (top-left, top-right, bottom-left, and bottom-right) along the center point, and we regress the building contours by leveraging the features of these sub-region center points. As depicted in Fig. \ref{fig:4}, compared to Fig. \ref{fig:4}(a) where contour vertex offsets are regressed solely from the center point, the sub-region center point in Fig. \ref{fig:4}(b) is closer to the building contour in relative coordinate position, Moreover, the coupled features of sub-region center points offer richer building contour information. 

Specifically, the feature corresponding to the positions of the sub-region center points are derived from the feature map \( F \) via bilinear interpolation. These feature vectors are then concatenated and fed into the offset prediction head, which consists of two stacked linear layers and ReLU activation functions, producing \( N \) contour vertex offsets \(\{(\Delta x_i, \Delta y_i) \mid i = 1, 2, \ldots, N\}\). The predicted offsets are subsequently added to the four sub-region center points to form the initial building contour.

To ensure alignment with the ground truth contour vertices during the training process, we employ the contour description methodology \cite{zhang2022e2ec}. This approach organizes the contour as a sequence of vertex coordinates arranged in either a counterclockwise or clockwise direction. Initially, four control points are inserted at four cardinal directions of the building (bottom, right, top, and left). The segments of the building contour between each pair of control points are then densely resampled to achieve an equal number of vertices. Consequently, a building contour consisting of \( N \) vertex coordinates is obtained, with an equal number of vertices between each pair of control points. In our experiments, following E2EC \cite{zhang2022e2ec}, $N$ is set to 128. 
% In our experiments, the ground truth building contour \({gt} = \{ p_i \}_{i=1}^N \) is set to 128 vertices.

The contour generation process follows the order of label creation, such as right-bottom, right-top, left-top, and left-bottom. Specifically, the contour vertices of the building's right-bottom part are generated by adding the right-bottom sub-region center point to the first 32 offsets, and so on to generate the initial contour of the building.
\begin{equation}
\begin{small}
\begin{aligned}
&{p}^{{rb}} = (x_i^{{rb}}, y_i^{{rb}}) + \lambda \times \{(\Delta x_i, \Delta y_i) \mid i = 1, \ldots, \frac{N}{4}\}, \\
&{p}^{{rt}} = (x_i^{{rt}}, y_i^{{rt}}) + \lambda \times \{(\Delta x_i, \Delta y_i) \mid i = \frac{N}{4} + 1, \ldots, \frac{N}{2}\}, \\
&{p}^{{lt}} = (x_i^{{lt}}, y_i^{{lt}}) + \lambda \times \{(\Delta x_i, \Delta y_i) \mid i = \frac{N}{2} + 1, \ldots, \frac{3N}{4}\}, \\
&{p}^{{lb}} = (x_i^{{lb}}, y_i^{{lb}}) + \lambda \times \{(\Delta x_i, \Delta y_i) \mid i = \frac{3N}{4} + 1, \ldots, N\}.
\end{aligned}
\end{small}
\end{equation}
$(x_i^{lt}, y_i^{lt})$, $(x_i^{lb}, y_i^{lb})$, $(x_i^{rt}, y_i^{rt})$, $(x_i^{rb}, y_i^{rb})$ represent the coordinates for the left-top, left-bottom, right-top, and right-bottom sub-region center points, $(\Delta x_i, \Delta y_i)$ represent offsets.

To address the significant variability in regression distances between sub-region center points and contour vertices across different buildings, we introduce a scaling factor $\lambda$, which is set to 10. $\lambda$ helps maintain model stability by reducing the impact of large variations in regression distances.
% To address the significant variability in regression distances between sub-region center points and contour vertices across different buildings, we introduce a scaling factor $\lambda$, which is set to 10. $\lambda$ helps maintain model stability by reducing the impact of large variations in regression distances.

The generation order of ground truth contour vertices is fixed, and we supervise the offset prediction head using the smooth L1 loss:

\begin{equation}
\label{deqn_ex1a}
\begin{array}{ll}
L_{init} = \frac{1}{N} \sum_{i=1}^{N} l_1 (py_{i}^{{init}} - gt_i)
\end{array}
\end{equation}
Where \( py_{_i}^{{init}} \) represents the initial contour vertex predicted by ICGM, \({gt}_i \) represents the ground truth building contour vertex, and \( l_1 \) denotes the smooth L1 loss function.

% where \(py_{i}^{{init}}\) represents the initial contour predicted by the ICGM, \(l_1\) represents the smooth L1 loss function.
% where \(py_{i}^{{init}}\) represents the initial contour predicted by the ICGM, \(l_1\) represents the smooth L1 loss function.

\subsection{Contour Optimization}
The initial building contours generated by the ICGM are often rough, requiring further refinement through a dedicated contour optimization. Contour optimization is designed to enhance contour precision through a two-tiered adjustment process: contour coarse adjustment and contour precise adjustment. The coarse adjustment phase utilizes concatenated features from the contour vertices and sub-region centers to perform a preliminary global adjustment of the contour vertices. The precise adjustment phase employs a Contour Optimization Module (COM) integrated CNN-Transformer architecture, which processes detailed contour feature descriptions from both global and local vertices. This architecture adapts dynamically to capture intricate dependencies between contour vertices, iterating three times to produce contour polygons.

\subsubsection{Contour Coarse Adjustment}
Given an initial contour, we first construct a feature vector \( F_i \) for each building contour by extracting the feature information of each contour vertex from the feature map \( F \) using bilinear interpolation. The features of the \( N \) vertices are then concatenated to generate the feature representation of the building contour, denoted as \( f_i \). Additionally, to enhance global information, we concatenate the center point features of the four sub-regions \( f_i^{ct} \) with \( f_i \) to form an \( (N+4) \times C \) feature vector (where \( C \) represents the number of feature channels), thereby improving control over the overall offset magnitude. Both \( f_i \) and \( f_i^{ct} \) are obtained through bilinear interpolation. This feature vector is fed into a multi-layer perceptron (MLP) composed of two hidden layers, outputting the offset predictions for each vertex \(\{(\Delta x_i^{co}, \Delta y_i^{co}) \mid i = 1, 2, \ldots, N\}\). The offsets are then added to the initial contour to update the vertex coordinates \((x_i^{co}, y_i^{co}) = (x_i, y_i) + \omega \times (\Delta x_i^{co}, \Delta y_i^{co})\)

In our experiments, \( C \) is set to 64. To ensure more stable regression of vertex offsets, we introduce a scaling factor $\omega$, which is set to 4. This helps control the range of offset values and improves the robustness of the model. Smooth L1 loss is used to supervise the contour coarse adjustment offset loss \( L_{{coarse}} \).

\begin{equation}
\label{deqn_ex1a}
\begin{array}{ll}
L_{{coarse}} = \frac{1}{N} \sum_{i=1}^{N} l_1 (py_{i}^{{co}} - gt_i)
\end{array}
\end{equation}
where \( py_{i}^{co} \) represents the coarse adjusted contours.

\subsubsection{Contour Optimization Module Integrated CNN-Transformer}
It is worth noting that the building contour is a set of coordinate vertices linked end to end. The vertex offset is not only affected by the feature of the current vertex, but also associated with its adjacent vertices. Therefore, it is necessary to query the associated vertices from the contour features to achieve accurate prediction of offset vectors. However, limited by the size of the convolution kernel, it is difficult for circular convolution to capture the global features of the contour. Therefore, we integrate CNN and Transformer to design a novel contour optimization structure, which can extract semantic information between global and local vertices, automatically capture feature dependencies between vertices at different positions, and iteratively output more accurate vertex offsets.
 
 \begin{figure*}[]
    \centering
    \includegraphics[width=0.6 \linewidth]{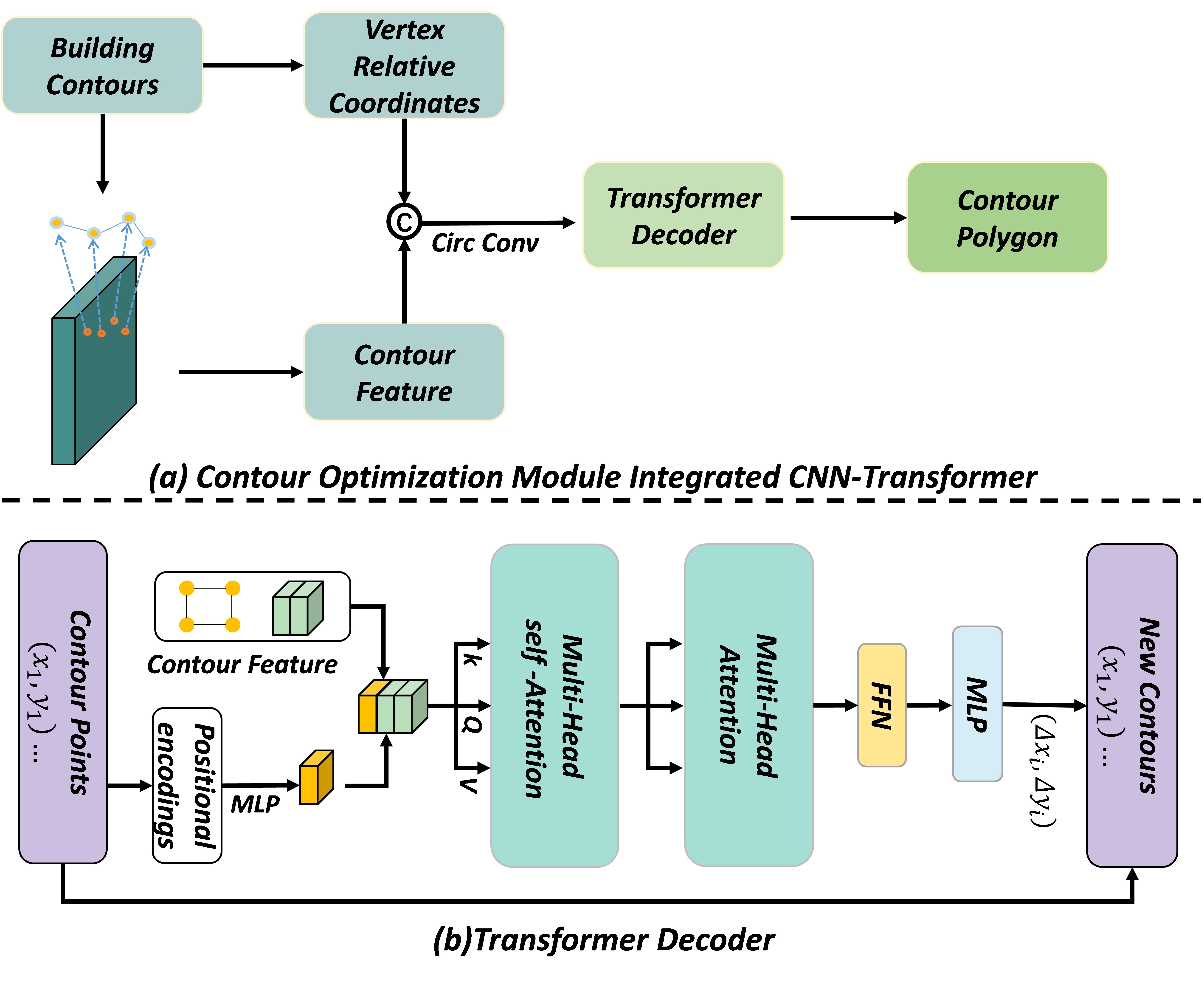}
    \vspace{-1em} % 根据需要调整这个值
    \caption{The architecture of contour optimization module integrated CNN-Transformer.}
    \label{fig:com}
\end{figure*}
The structure of the contour optimization module is shown in Fig. \ref{fig:com}(a). First, the features of \( N \) contour vertices are combined with the vertex relative positions \((x_i^{rel}, y_i^{rel})\) to obtain a feature vector \( F_i^{cont} \) with a length of \( (C+2) \times N \) (where \( C \) is the channel number of the feature). Then, \( F_i^{cont} \) is input into the circular convolution to obtain the feature description \( F_i \) of the building contour. Finally, a Transformer decoder is employed to automatically capture the dependencies between vertices at different positions of the building contour features \( F_i \). By extracting both local and global features, the Transformer decoder precisely predicts vertex offsets. These offsets are then added to the building contours from the previous iteration to complete an optimization step.

Fig. \ref{fig:com}(b) illustrates the interior of the Transformer decoder. Initially, the current contour vertex coordinates undergo position encoding and are input into a multi-layer perceptron (MLP) to map the positional information's features. This feature mapping is then concatenated with the contour feature representation \( F_i \). This approach effectively reduces the network's learning complexity while enhancing contour extraction performance. Subsequently, multi-head self-attention and multi-head attention mechanisms capture dependencies and feature correlations among contour vertices. Each decoder output is processed through an MLP comprising two stacked linear layers followed by a ReLU activation function. This MLP generates precise vertex offsets \(\{(\Delta x_i^{it}, \Delta y_i^{it}) \mid i = 1, 2, \ldots, N\}\). During each iteration, the contour vertex coordinates are updated as:\((x_i^{it}, y_i^{it}) = (x_i, y_i) + (\Delta x_i^{it}, \Delta y_i^{it})\)

The contour optimization module optimizes the building contours through three iterations, so it is necessary to supervise the contours optimized in each iteration. However, the smooth L1 loss has the problem of over-smoothing. Therefore, we use the Smooth L1 loss to calculate the losses \( L_{it1} \) and \( L_{it2} \) of the first and second iteration optimizations, and use DML \cite{zhang2022e2ec} to calculate the loss of the third optimization \( L_{it3} \).
\begin{equation}
\label{deqn_ex1a}
\begin{array}{ll}
L_{{it1}} = \frac{1}{N} \sum_{i=1}^{N} l_1 (py_{i}^{{it1}} - gt_i)
\end{array}
\end{equation}
\begin{equation}
\label{deqn_ex1a}
\begin{array}{ll}
L_{it2} = \frac{1}{N} \sum_{i=1}^{N} l_1 (py_{i}^{it2} - gt_i)
\end{array}
\end{equation}
where \(py_{i}^{it1}\) and \(py_{i}^{it2}\) respectively represent the contours after the first and second optimization.
% \minew{DML consists of two parts. 1) the distance between the predicted vertex points (orange) and the nearest point on the label boundary, which is represented by 10 times densified points (green) for each segment, denoted as the black arrow in Fig. \ref{dlm}(a); and 2) the distance between each corner vertex (purple) and its nearest predicted vertex, as shown in Fig. \ref{dlm}(b). 

\begin{figure}[]
    \centering
    \includegraphics[width=9cm]{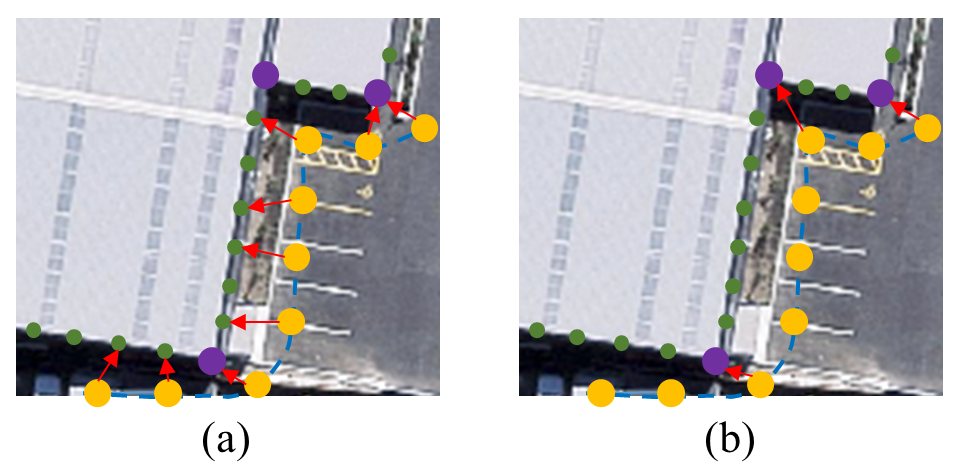}
    \vspace{-2.5em} % 根据需要调整这个值
    \caption{Example of dynamic matching loss. (a) A predicted contour vertex is pulled toward the nearest point on the densified label boundary. (b) A building corner vertex pulls the nearest predicted vertex toward its location. Orange: predicted contour vertices. Purple: corner vertices of the ground-truth polygon. Green: vertices of the 10 × densified ground truth.}
    \label{dlm}
\end{figure}

% Firstly, equation (10) assigns each predicted contour vertex $\hat{p}_i$ to the nearest interpolated ground-truth (gt) contour vertex $gt_{x_i^p}$. Equation (11) calculates the L1 norm loss between each predicted contour node and the matched gt contour vertex $gt_{(x_i^*)^p}$. Equation (12) assigns each gt contour corner point $gt_i^{key}$ to the nearest predicted contour vertex $\hat{p}_{(y_i^*)}$ using the minimum Euclidean distance. Then, equation (13) calculates the L1 norm loss between each gt contour corner point $gt_i^{key}$ and the matched predicted contour vertex $\hat{p}_{(y_i^*)}$. The final DML loss is composed of $L_{p1}$ and $L_{p2}$.
% }
Dynamic Matching Loss (DML) consists of two parts. 1) the distance between the predicted vertices (orange) and the nearest ground truth contour vertex on the label boundary, which is represented by 10 times densified points (green) for each segment, denoted as the red arrow in Fig. \ref{dlm}(a); and 2) the distance between each corner vertex (purple) and its nearest predicted vertex, as shown in Fig. \ref{dlm}(b).

Firstly, equation (10) assigns each predicted vertex $\hat{p}_i$ to the nearest interpolated ground truth contour vertex ${gt}_x$. Equation (11) calculates the L1 norm loss between each $\hat{p}_i$ and its matched ground truth vertex ${gt}_x$. Equation (12) calculates the minimum Euclidean distance between each ground truth corner vertex $gt_i^{key}$ and the predicted vertex $\hat{p}_y$ to assigns each $gt_i^{key}$ to the nearest predicted vertex $\hat{p}_{y_i^*}$, and equation (13) calculates the L1 norm loss between each $gt_i^{key}$ and the matched predicted vertex $\hat{p}_{y_i^*}$. The final DML loss combines $L_{p1}$ and $L_{p2}$ to optimize predicted vertex alignment with the ground truth boundary and corners. For more details, please refer to E2EC.

%First, the interpolated ground truth is marked as \( {gt}^{ip} \), and the corresponding nearest interpolated ground truth contour vertex is found for each predicted contour vertex through Eq (10). The first loss of DML is calculated by Eq (11). The second loss of DML aims to recover the contour polygon. The ground truth key vertex is dynamically matched with the closest predicted contour vertex by Eq (12), and the second loss is calculated by Eq (13). The DML is the sum of two parts.

\begin{equation}
\label{deqn_ex1a}
\begin{array}{ll}
x_i^* = argmin \| \hat{p}_i - {gt}_x \|_2
\end{array}
\end{equation}
\begin{equation}
\label{deqn_ex1a}
\begin{array}{ll}
L_{p1}({pre}, {gt}) = \frac{1}{N} \sum_{i=1}^{N} \| \hat{p}_i - {gt}_{x_i^*} \|_1
\end{array}
\end{equation}
\begin{equation}
\label{deqn_ex1a}
\begin{array}{ll}
y_i^* = argmin \| \hat{p}_y - {gt}_i^{key} \|_2
\end{array}
\end{equation}
\begin{equation}
\label{deqn_ex1a}
\begin{array}{ll}
L_{p2}({pre}, {gt}) = \frac{1}{N_{key}} \sum_{i=1}^{N_{key}} \| \hat{p}_{y_i^*} - {gt}_i^{key} \|_1
\end{array}
\end{equation}
\begin{equation}
\label{deqn_ex1a}
\begin{array}{ll}
L_{it3} = \frac{L_{p1} + L_{p2}}{2}
\end{array}
\end{equation}
where \({gt}_i^{key}\) represents the original corner vertices of ground truth building contour, as shown by the purple vertex in Fig. \ref{dlm}.

\section{Experimental Setting}
\subsection{Datasets}
To evaluate the performance of PolyBuild, we use two publicly available building datasets: the WHU aerial building dataset \cite{ji2018fully} , the WHU-Mix (Vector) building dataset
\cite{wei2023buildmapper} and Crowd AI dataset \cite{mohanty2018crowdai}.
\subsubsection{WHU aerial building dataset}
The WHU aerial building dataset contains a large number of high-resolution aerial images and corresponding accurate building labels. This aerial dataset has a resolution of 0.2 meters and includes 187,000 buildings of various types and uses. All aerial images are cropped into 512 × 512 tiles. The training set, validation set, and test set consist of 10,961, 2,486, and 6,177 tiles, respectively.
\subsubsection{WHU-Mix (Vector) building dataset}
% \minew{Table \ref{whu-mix} lists the detailed information of WHU-Mix (vector) building dataset.} 

The WHU-Mix (Vector) building dataset, hereinafter referred to as the WHU-Mix dataset, is composed of multiple datasets, including Crow AI, Open AI \cite{yap2018open}, SpaceNet \cite{van2018spacenet}, Inria  \cite{maggiori2017can}, WHU datasets \cite{ji2018fully}, and new data. In the WHU-Mix dataset, the Crow AI and SpaceNet sub-datasets are satellite images. These datasets exhibit significant differences in building styles and geographic environments, allowing for a comprehensive evaluation of the robustness of various methods. The WHU-Mix dataset consists of over 64,000 tile images, representing more than 754,000 individual buildings and covering approximately 1,100 square kilometers of total geographic area. The WHU-Mix Dataset is divided into four distinct subsets: the training set, validation set, test set I, and test set II, comprising 43,778, 2,922, 11,675, and 6,011 tiles, respectively. It is worth noting that test set II does not overlap geographically with the training set, allowing for further validation of the methods' generalization capabilities.

\subsubsection{Crowd AI dataset}
Crowd AI provides over 340,000 individual satellite imagery tiles, each with a size of 300 × 300 pixel RGB images and building annotations. The training set and test set consist of 280,741 and 60,317 tiles, respectively, along with their corresponding annotations in MS-COCO format. Due to the large sample size, Crowd AI requires higher computational costs, but the samples lack diversity.

% \begin{table*}[]
% \centering
% {\setlength{\tabcolsep}{5mm}
% \caption{\minew{The details of the WHU-Mix (vector) building dataset. In the WHU-Mix dataset, the Crow AI and SpaceNet sub-datasets are satellite images.}}
% \label{whu-mix}
% \begin{tabular}{lcccccc}
%        \toprule
%         No. & Data Source & Region & Resolution (m) & Size (pixel) & Tile quantity & Building quantity \\
%         \midrule
%         1 & Crowd AI &-  & 0.20 & 300 & 15,000 & 129,104 \\
%         2 & SpaceNet & Vegas, Khartoum & 0.30 & 650 & 4,521 & 132,491 \\
%         3 & Open AI & Tanzania & 0.08 & 512 & 7,246 & 23,096 \\
%         4 & WHU & Christchurch & 0.20 & 512 & 14,531 & 237,674 \\
%         5 & WHU & East Asia & 0.34 & 512 & 5,618 & 40,792 \\
%         6 & New & Zhejiang & 0.20 & 512 & 6,009 & 40,183 \\
%         7 & New & Hubei & 0.64 & 512 & 3,025 & 66,445 \\
%         8 & New & Chongqing & 0.15 & 512 & 2,426 & 5,986 \\
%         9 & Inria & Tyrol, Kitsap & 0.30 & 512 & 2,971 & 35,311 \\
%         10 & SpaceNet & Paris & 0.30 & 650 & 633 & 16,121 \\
%         11 & New & Wixi & 0.17 & 512 & 1,267 & 6,305 \\
%         12 & New & Dunedin & 0.20 & 512 & 1,140 & 20,618 \\
%         Total Quantity &  &  &  &  & 64,387 & 754,126 \\
% \bottomrule
% \end{tabular}}
% \end{table*}

\subsection{Experimental details}
The proposed framework is trained using the PyTorch library on an RTX 2080ti GPU. It undergoes 150 epochs of training on the WHU aerial dataset, employing the Adam optimizer with an initial learning rate set to \( 1 \times 10^{-4} \), decayed at epochs 80 and 120. Additionally, it is trained on the WHU-Mix dataset for 30 epochs, starting with a learning rate of \( 1 \times 10^{-4} \), decayed at epochs 15 and 20, also utilizing the Adam optimizer for parameter updates.

PolyBuild jointly optimizes multiple tasks, incorporating a comprehensive loss function that includes several components: the building center point heat map loss \(L_{ht}\), the building size loss \(L_{wh}\), the initial contour loss \(L_{init}\), the contour coarse adjustment loss \(L_{\text{coarse}}\), and the losses for the three contour optimization iterations \(L_{it1}\), \(L_{it2}\), and \(L_{it3}\). The joint loss function we define is therefore:
\begin{equation}
\begin{split}
L = & L_{ht} + (L_{wh} + L_{init} + L_{coarse}) \times \phi \\
    & + (L_{it1} + L_{it2} + L_{it3}) \times \tau
\end{split}
\end{equation}
where \( \phi \) and \( \tau \) are set to 0.1 and 1/3, respectively.

% PolyBuild jointly optimizes multiple tasks, and the total loss includes the building center heatmap loss \( L_{ht} \), building width and height loss \( L_{wh} \), initial contour loss \( L_{int} \), coarse contour adjustment loss \( L_{coarse} \), first accurate contour adjustment loss \( L_{ite1} \), second accurate contour adjustment loss \( L_{ite2} \), and the third accurate adjustment loss \( L_{dml} \). It is worth noting that the third fine adjustment loss uses DML loss \cite{24}, while other contour losses use the smooth L1 loss. Therefore, our defined loss function is:
% \begin{equation}
% \label{deqn_ex1a}
% \text{Smooth L1 Loss}(x, y) = \begin{cases} 
% 0.5 \cdot (x - y)^2 & \text{if } |x - y| < \beta \\
% |x - y| - 0.5 \cdot \beta & \text{otherwise} 
% \end{cases} 
% \end{equation}

\subsection{Evaluation Metrics}
We adopt standard COCO evaluation methods to assess the instance-level building extraction performance, including average precision \( AP \) and average recall \( AR \) at different thresholds. Intersection over union (IoU) is the ratio of the intersection area to the union area of the predicted and actual building areas. \( AP \) is the average precision at 10 different IoU values (from 0.5 to 0.95, with a step size of 0.05), and \( AR \) is calculated similarly to \( AP \). We additionally report the average precision at IoU thresholds of 0.5 and 0.75, denoted as \( AP_{50} \) and \( AP_{75} \).
\begin{equation}
\label{deqn_ex1a}
IoU(G, P) = \frac{{pre} \cap {gt}}{{pre} \cup {gt}}
\end{equation}
\begin{equation}
\label{deqn_ex1a}
{AP} = \frac{{AP}_{0.50} + {AP}_{0.55} + \dots + {AP}_{0.95}}{10} 
\end{equation}
\section{Results and discussion}
\subsection{Comparison with the state-of-the-art method}
In this section, we conduct a comparative analysis of PolyBuild against advanced mask-based instance segmentation methods (Mask R-CNN \cite{he2017mask}, SOLO \cite{wang2020solo}, YOLACT \cite{bolya2019yolact}, RTMDet \cite{lyu2022rtmdet}) and contour-based instance segmentation methods (DeepSnake \cite{peng2020deep}, CLP-CNN \cite{wei2021concentric}, E2EC \cite{zhang2022e2ec}, Hisup \cite{xu2023hisup}, BuildMapper \cite{wei2023buildmapper}, Line2Poly \cite{wei2024lines}).

\begin{table*}[]
\centering
{\setlength{\tabcolsep}{3mm}
\caption{Quantitative results of different methods on the WHU aerial building dataset. The table highlights the highest values in \textcolor{red}{red}, and the second-highest results in \textcolor{blue}{blue}.}
\label{1}
\begin{tabular}{lccccc}
\toprule
Method& Backbone& \( AP \) (\%) & \( AP_{50} \) (\%) & \( AP_{75} \) (\%) & \( AR \) (\%) \\
\midrule
\textit{Mask-based}:& ~ & ~ & ~ & ~ & ~ \\
Mask R-CNN \cite{he2017mask}& ResNet50& 65.3 & 90.0 & 77.1 & 70.7 \\
YOLACT \cite{bolya2019yolact}& ResNet50& 65.3 & 88.5 & 76.5 & 71.1 \\
SOLO \cite{wang2020solo}& ResNet50& 68.6 & 89.8 & 79.4 & 73.4 \\
RTMDet \cite{lyu2022rtmdet}& ResNet50& 69.6 & 88.5 & 79.5 & 77.7 \\
\textit{Contour-based}:  & ~ & ~ & ~ & ~ & ~ \\
Deep Snake \cite{peng2020deep}&DLA-34  & 72.7 & 91.5 & 82.8 & 78.6 \\
CLP-CNN \cite{wei2021concentric}&DLA-34  & 72.6 & 90.9 & 82.6 & 78.0 \\
E2EC \cite{zhang2022e2ec}&DLA-34  & 72.0 & 89.8 & 80.9 & 80.5 \\
Hisup \cite{xu2023hisup}& HRNetV2-W48 & 73.5 & 86.7 &  79.8 & - \\
BuildMapper \cite{wei2023buildmapper}& DLA-34 & 73.6  & 89.0   & 81.6   & 78.9 \\
Line2Poly \cite{wei2024lines}& DLA-34 & \textcolor{blue}{73.8}  & \textcolor{blue}{89.4}   & \textcolor{blue}{82.1}   & \textcolor{blue}{79.7}   \\
PolyBuild & DLA-34 &\textcolor{red}{74.0} & \textcolor{red}{91.1} &\textcolor{red} {82.5} & \textcolor{red}{80.7} \\
\bottomrule
\end{tabular}}
\end{table*}
Table \ref{1} presents the quantitative comparison of different methods on the WHU aerial building dataset. Contour-based methods significantly outperform mask-based methods, highlighting the advantages of contour-based approaches in building extraction. PolyBuild stands out with the highest \( AP \) and \( AR \), reaching 74.0\% and 80.7\%, respectively. The second place is Line2Poly, which integrates CNN and DETR \cite{carion2020end} architectures to achieve excellent contour extraction; however, the contours generated based on feature lines are overly complex. Hisup enhances vertex extraction accuracy by incorporating segmentation maps, resulting in an average precision (AP) of 73.5\%. Deep Snake produces initial contours in the form of octagons, which complicates the subsequent contour optimization process. Compared to PolyBuild, its \( AP \) and \( AR \) are reduced by 1.3\% and 2.1\%, respectively. CLP-CNN competes closely with DeepSnake. Although CLP-CNN is specifically designed for building extraction, its concentric loop convolution increases the network parameters and the redundancy of some features affects the contour extraction accuracy. PolyBuild improves \( AP \) and \( AR \) indicators by 8.7\% and 10\% over the classic Mask R-CNN and by 4.7\% and 7.2\% over SOLO. While Deepsnake, CLP-CNN, and E2EC generate building vector contours end-to-end, they employ fixed-size recurrent circular convolutions for predicting contour vertex offsets during evolution, which limits model accuracy. In contrast, PolyBuild integrates CNN and transformer architectures to adaptively capture dependencies among contour vertex features, achieving superior accuracy in contour polygon extraction through iterative position encoding updates.

 \begin{figure*}[]
    \centering
    \includegraphics[width=18cm]{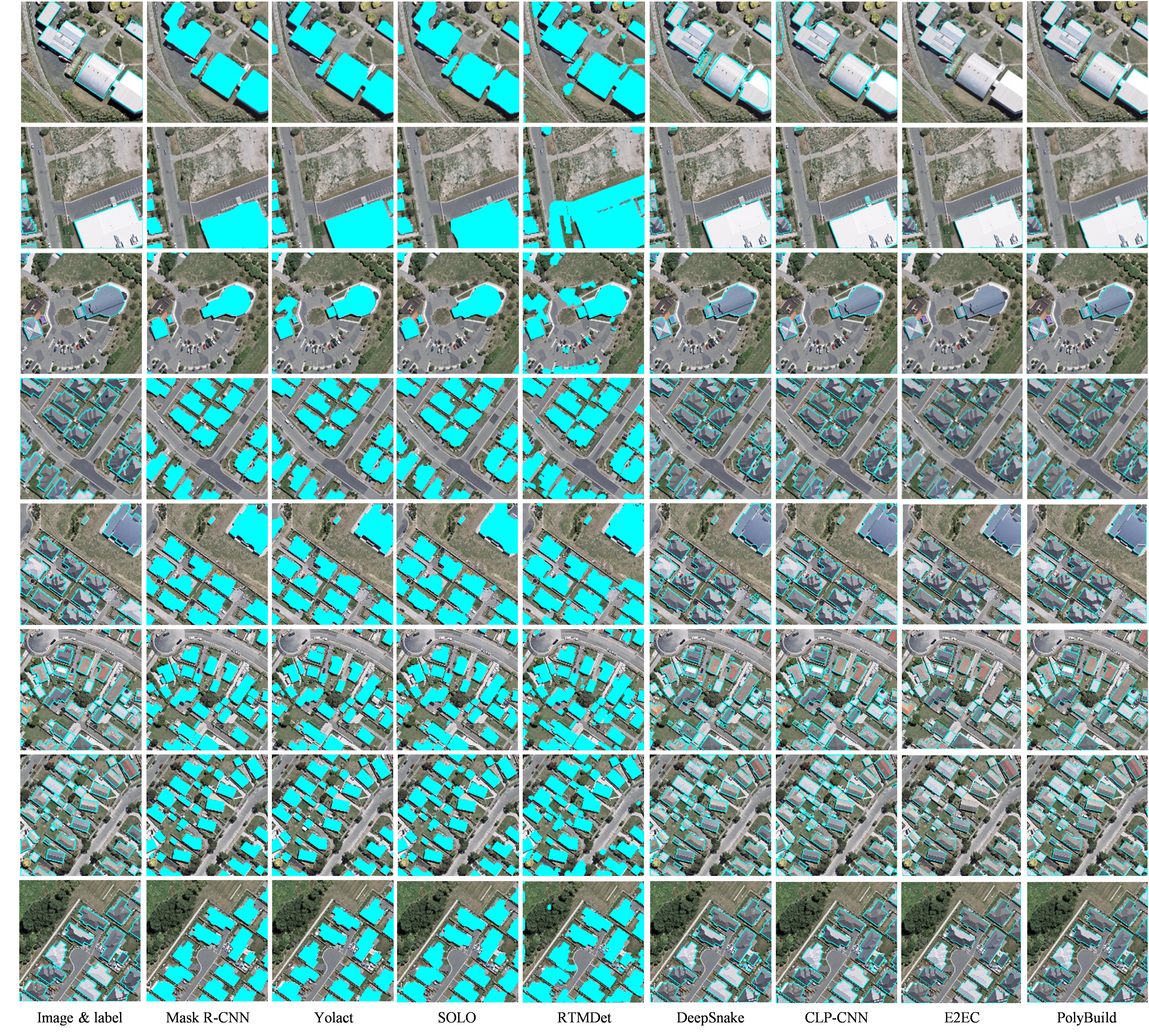}
    \vspace{-1em} % 根据需要调整这个值
    \caption{Visualization results of different methods on the WHU aerial building dataset.}
    \label{fig:WHU}
\end{figure*}
\begin{table*}[]
\centering
{\setlength{\tabcolsep}{3mm}
\caption{Quantitative results of different methods on the WHU-Mix building dataset. The table highlights the highest values in \textcolor{red}{red}, and the second-highest results in \textcolor{blue}{blue}.}
\label{2}
\begin{tabular}{lccccccccccc}
\toprule
Method & Backbone & \multicolumn{4}{c}{WHU-Mix test set I (\%)} & \multicolumn{4}{c}{WHU-Mix test set II (\%)} & \multicolumn{2}{c}{Test set I \& II (\%)} \\

& & \( AP \) & \( AP_{50} \) & \( AP_{75} \) & \( AR \) & \( AP \) & \( AP_{50} \) & \( AP_{75} \) & \( AR \) & \( mAP \) & \( mAR \) \\
\midrule
\textit{Mask-based}: & ~ & ~ & ~ & ~ & ~ & ~ & ~ & ~ & ~ & ~ & ~\\
Mask R-CNN \cite{he2017mask}&ResNet50& 47.0 & 67.0 & 53.2 & 53.7 & 46.1 & 73.9 & 49.0 & 54.8 & 46.5 & 54.3 \\
YOLACT \cite{bolya2019yolact}&ResNet50& 42.3 & 65.7 & 47.2 & 49.7 & 41.3 & 71.3 & 42.3 & 50.6 & 41.8 & 50.2 \\
SOLO   \cite{wang2020solo} &ResNet50& \textcolor{blue}{57.1} & {83.2} & \textcolor{red}{65.1} & \textcolor{blue}{63.9} & 45.3 & 74.3 & 47.9 & 54.7 & 51.2 & 59.3 \\
RTMDet  \cite{lyu2022rtmdet} &ResNet50& 43.1 & 65.3 & 47.9 & 59.4 & 45.0 & 72.5 & 47.7 & \textcolor{red}{59.8} & 45.5 & 59.6 \\
Mask2Fromer \cite{cheng2022masked} &ResNet50&  \textcolor{red}{57.6} &  \textcolor{red}{84.9} &  64.6 & - & 48.2 &  \textcolor{red}{78.1} &  51.1 & - & \textcolor{blue}{52.9} & -\\
\textit{Contour-based}: & ~ & ~ & ~ & ~ & ~ & ~ & ~ & ~ & ~ & ~\\
PolarMask \cite{xie2020polarmask} &ResNet50  &44.8 & 69.1 &  50.7 & - &  39.1 & 66.3& 40.5 & - & 42 & - \\

Deep Snake  \cite{peng2020deep}&DLA-34& 55.3 & 82.1 & 63.0 & 61.8 & 46.9 & 73.9 & 51.5 & 54.8 & 51.1 & 58.3 \\
CLP-CNN  \cite{wei2021concentric}&DLA-34& 55.6 & 81.8 & {63.5} & 62.3 & 48.3 & 75.2 & 52.4 & 56.4 & 52.0 & 59.4 \\
E2EC \cite{zhang2022e2ec}&DLA-34& 56.4 & {83.2} & 62.7 &63.3 & \textcolor{blue}{48.9} & {75.4} & \textcolor{blue}{52.5} & 58.9 & 52.6 & \textcolor{blue}{61.1} \\
PolyBuild &DLA-34& \textcolor{red}{57.6} & \textcolor{blue}{83.3} & \textcolor{blue}{64.7} &  \textcolor{red}{65.2} &\textcolor{red}{49.4} & \textcolor{blue}{76.6 }& \textcolor{red}{53.1} & \textcolor{blue}{59.1} & \textcolor{red}{53.5} & \textcolor{red}{62.2} \\
\bottomrule
\end{tabular}}
\end{table*}

% \minew{We also report the performance of different methods and PolyBuild on the Crowd AI dataset in Table \ref{crowd}. PolyBuild shows advanced performance. Mask R-CNN \cite{he2017mask}, PANet \cite{liu2018path}, and HTC-DP \cite{zhao2020building} are all mask-based methods with poor accuracy compared to PolyBuild. Building Outline Delineation (BOD) \cite{liu2022building} and PolyMapper \cite{li2019topological} methods use RNN to iteratively predict building corner vertices, which is prone to missing vertices, resulting in relatively poor accuracy.}

\begin{table}[ht]
\centering
{\setlength{\tabcolsep}{2.5mm}
\caption{Quantitative results of different methods on the Crowd AI dataset.}
\label{crowd}
\begin{tabular}{lcccc}
\hline
{Method}  & \( AP \) (\%) & \( AP_{50} \) (\%) & \( AP_{75} \) (\%) & \( AR \) (\%) \\
\hline
Mask R-CNN \cite{he2017mask} & 41.9 & 67.5 & 48.8 & 47.6 \\
HTC-DP \cite{zhao2020building} & 44.4 & - & - & - \\
BOD \cite{liu2022building}& 47.4 & - & - & - \\
PANet \cite{liu2018path}& 50.7 & 73.9 & 62.6 & 54.4 \\
PolyMapper \cite{li2019topological}& 55.7 & 86.0 & 65.1 & 62.1 \\
PolyBuild & 60.6 & 87.8 & 69.3 & 63.9 \\
\hline
\end{tabular}}

\end{table}

To assess the generalization capability and robustness of PolyBuild, we conducted comparative experiments with various methods on the WHU-Mix building dataset. Table \ref{2} presents the evaluation results of different methods on the WHU-Mix dataset, demonstrating that PolyBuild maintains its leading position. PolyBuild achieved 53.5 \( mAP \) and 62.2 \( mAR \) on the WHU-Mix dataset, surpassing the mask-based state-of-the-art (SOTA) by 0.6 \( AP \) and 2.6 \( AR \), and the contour-based SOTA by 0.9 \( mAP \).

The WHU-Mix test set II, which has no geographic overlap with the training set, thoroughly validates the generalization capability of the PolyBuild method. Compared to other methods, PolyBuild consistently maintains high extraction accuracy. Notably, PolyBuild employs the same object detector as E2EC, with its accuracy improvement attributed to the innovative initial contour generation module and the integrated CNN-Transformer-based contour optimization module. The initial contour generation module generates more stable and accurate initial contours based on sub-region center points, while the contour optimization module integrates a CNN-Transformer architecture to extract global and local features, iteratively refining the initial contours until producing regular polygonal contours.

We also report the performance of different methods and PolyBuild on the Crowd AI dataset in Table \ref{crowd}. PolyBuild shows advanced performance. Mask R-CNN \cite{he2017mask}, PANet \cite{liu2018path}, and HTC-DP \cite{zhao2020building} are all mask-based methods with poor accuracy compared to PolyBuild. These methods rely on pixel-level segmentation, which often struggles with precise boundary delineation, especially for complex building shapes. Building Outline Delineation (BOD) \cite{liu2022building} and PolyMapper \cite{li2019topological} methods use RNN to iteratively predict building corner vertices, which is prone to missing vertices, resulting in relatively poor accuracy. In contrast, PolyBuild leverages a combination of initial contour generation, coarse adjustment, and Transformer-based optimization to directly predict vectorized contours, effectively addressing the limitations of mask-based and RNN-based approaches. This design enables PolyBuild to achieve more accurate and consistent results.

 \begin{figure*}[]
    \centering
    \includegraphics[width=18cm]{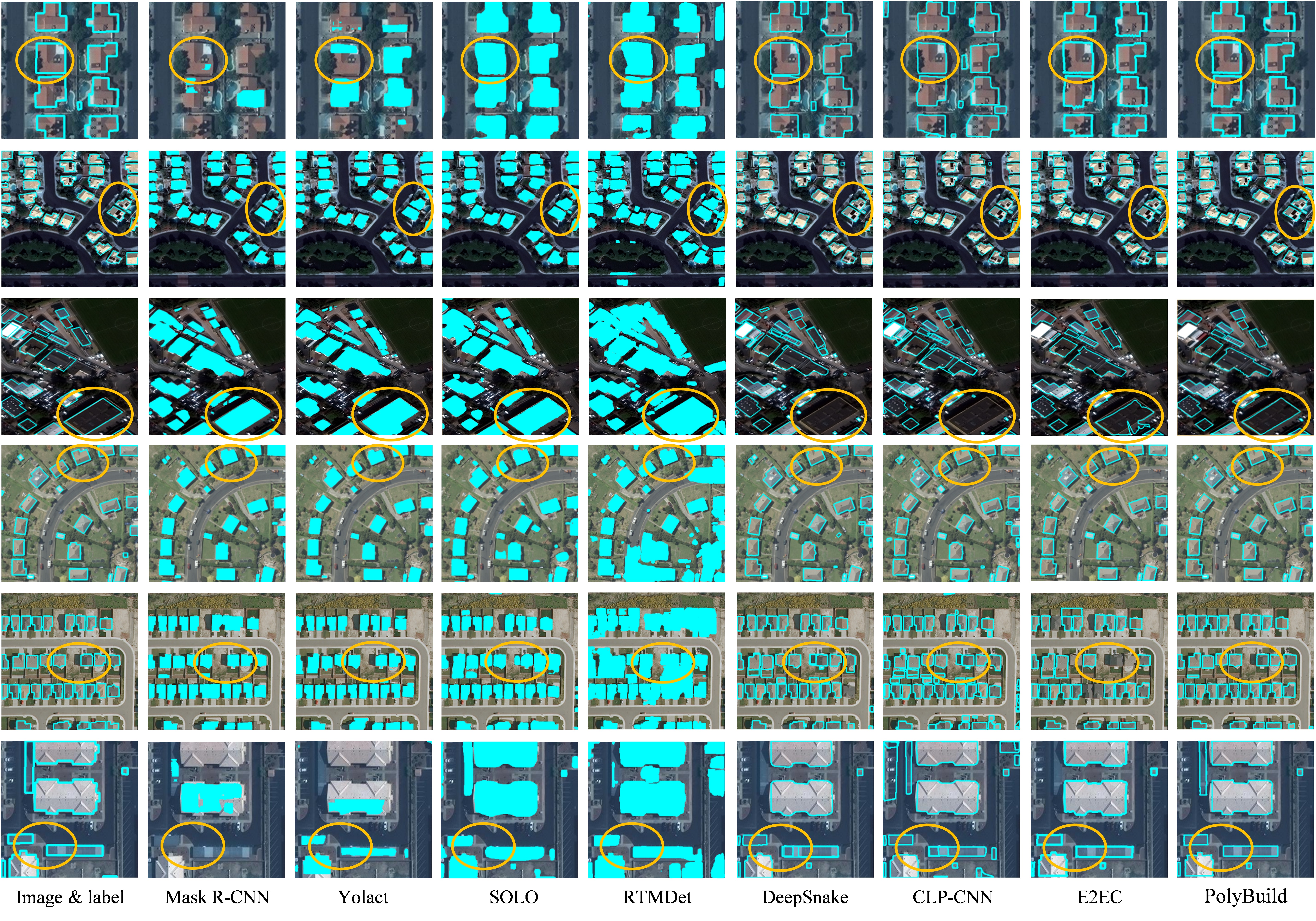}
    \vspace{-1em} % 根据需要调整这个值
    \caption{Visualization results of different methods on the WHU-Mix building dataset.}
    \label{fig:MIX}
\end{figure*}
Fig. \ref{fig:WHU} and \ref{fig:MIX} present the visual results of different methods on the WHU aerial dataset and the WHU-Mix dataset. Compared to the WHU aerial dataset, the WHU-Mix dataset features a more complex background and incorporates multiple architectural styles, which increases the difficulty for network recognition and extraction.

From the figures, it is evident that the contours of buildings extracted using mask-based methods are blurry and fragmented, whereas contour-based methods can directly generate  contours of buildings. However, methods such as DeepSnake, CLP-CNN, and E2EC exhibit varying degrees of building omission. PolyBuild, on the other hand, incorporates an additional building bbox detector, enhancing the network's ability to supervise building recognition.

In the first three rows of Fig. \ref{fig:WHU}, other contour-based methods exhibit significant boundary errors for large buildings. In contrast, PolyBuild, by utilizing features from the center points of four sub-regions to generate the initial contour, significantly controls the error margin, thereby reducing the pressure on subsequent contour optimization. In the third row of Fig. \ref{fig:MIX}, the large building boundaries extracted by E2EC show substantial errors, with considerable coordinate changes between adjacent points leading to increased boundary errors. PolyBuild’s contour optimization module, designed with a CNN-Transformer structure, effectively accounts for the interdependencies among contour vertices, leveraging both global and local information for precise contour regression. For the extraction of small buildings, as shown in the fourth row of Fig. \ref{fig:MIX}, the results obtained by PolyBuild are comparable to the ground truth labels.
\subsection{Ablation study}

\begin{table}[]
\centering
\setlength{\tabcolsep}{1mm}
\caption{Comparison of initial contour accuracy generated by different strategies.}
\label{3}
\begin{tabular}{llcccc}
\toprule
Contour & \multicolumn{1}{c}{Strategy} & \( AP \) (\%) & \( AP_{50} \) (\%) & \( AP_{75} \) (\%) &  \( AR \) (\%) \\
\midrule
\multirow{4}{*}{\textit{Initial Contour}} & \multicolumn{1}{c}{A} & 64.7 & 88.4 & 75.4 & 70.1 \\
                                 & \multicolumn{1}{c}{B} & 64.7 & 88.6 & 75.6 & 70.2 \\
                                 & \multicolumn{1}{c}{C} & 64.6 & 88.2 & 74.9 & 70.2 \\
                                 & \multicolumn{1}{c}{D} & 66.9 & 88.8 & 77.0 & 72.1 \\
\bottomrule
\end{tabular}
\end{table}
In this section, we first evaluate different initial contour generation strategies on the WHU aerial building dataset, and then assess the performance of various modules in PolyBuild.

\begin{figure}[]
    \centering
    \includegraphics[width=9cm]{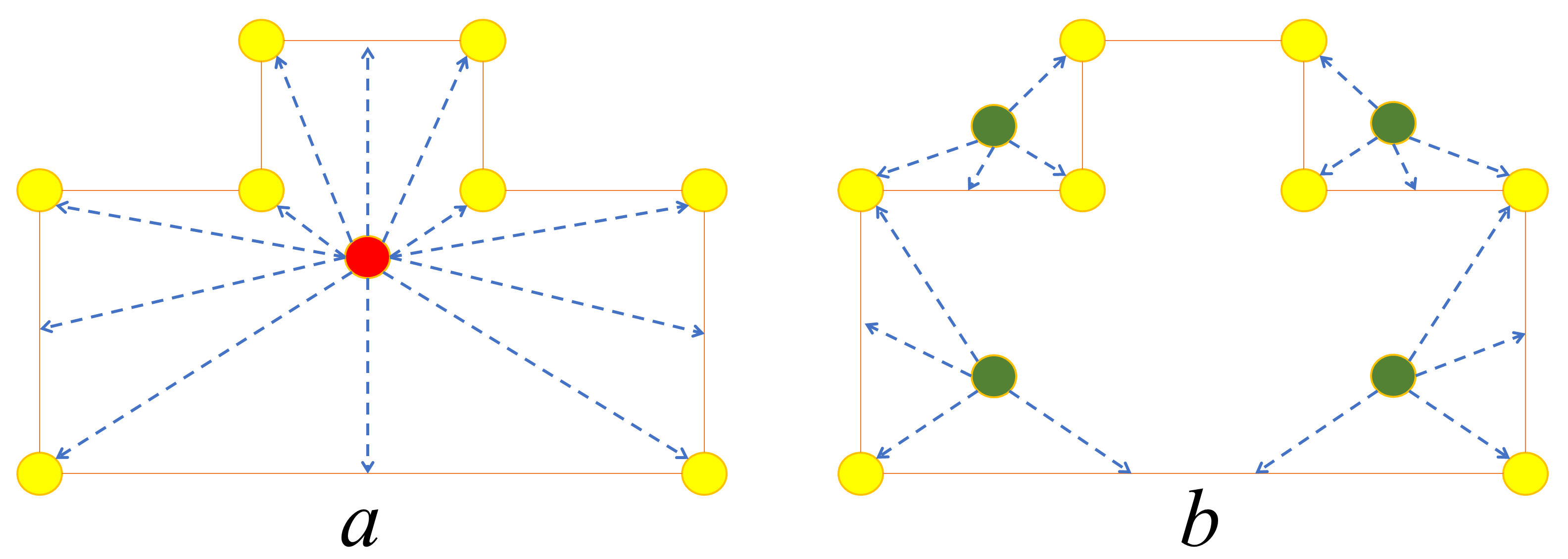}
    \vspace{-2em} % 根据需要调整这个值
    \caption{Visualization of initial contour generation methods under different strategies.}
    \label{fig:stage}
\end{figure}

We evaluate the accuracy of initial contours using different strategies. Strategy A involves predicting \( N \) offsets by concatenating sub-region center features and adding them to the center points to generate the initial contours. Strategy B independently predicts \( N/4 \) offsets for each sub-region center feature and adds them to the center points. Strategy C independently predicts \( N/4 \) offsets for each sub-region center point feature and adds them to the corresponding sub-region center points. Strategy D concatenates sub-region center features to predict \( N \) offsets, which are then added to the sub-region center points to generate the initial contours. Strategies A and B are depicted in Fig. \ref{fig:stage}(a), while strategies C and D are illustrated in Fig. \ref{fig:stage}(b).

As shown in Table \ref{3}, Strategy D achieves the highest accuracy in generating initial contours. Strategies A and B regress contour offsets based on sub-region center points but ultimately add these offsets to the center points, thereby complicating the contour generation process. Fig. \ref{fig:stage} illustrates that some sub-region center points are positioned outside the buildings, causing their corresponding feature information to inadequately capture the relevant building contours. Consequently, the accuracy of building contours generated by Strategy C, which relies on regressing offsets based on individual sub-region center features, is suboptimal. PolyBuild employs Strategy D, concatenating features from four sub-region centers to regress contour offsets and subsequently adding these offsets to the corresponding sub-region center points to generate the initial contours. This method significantly enhances both \( AP \) and \( AR \) compared to other strategies.

We also evaluate the accuracy of the initial building contour and the final output contour under different component combinations, using E2EC as the baseline network. The quantitative results are shown in Table \ref{4}.

For the initial contours, replacing the original network's contour initialization module with ICGM resulted in increases of 1.7 in  \( AP \) and 1.1 in \( AR \).  Although \( AP_{50} \) unchanged, \( AP_{75} \) increased by 1.4, indicating that ICGM generates more accurate initial outlines through more comprehensive regional center features.

\begin{table}[]
\centering
{\setlength{\tabcolsep}{1mm}
\caption{Ablation study of different module on the WHU aerial dataset.}
\label{4}
\begin{tabular}{lccccccc}
\toprule
Contour & ICGM & COM & \( AP \) (\%) & \( AP_{50} \) (\%) & \( AP_{75} \) (\%) & \( AR \) (\%) \\
\midrule
\multirow{2}{*}{\textit{Initial Contour}} & & & 65.2 & 88.8 & 75.6 & 71.0 \\
                                 & \checkmark & & 66.9 & 88.8 & 77.0 & 72.1 \\
\midrule
\multirow{3}{*}{\textit{Final Contour}} & & & 72.0 & 89.8 & 80.9 & 80.5 \\
                               & \checkmark & & 72.8 & 90.3 & 81.5 & 80.7 \\
                               & \checkmark & \checkmark & 74.0 & 91.1 & 82.5 & 80.7 \\
\bottomrule
\end{tabular}}
\end{table}
For the final output contours, replacing E2EC's contour initialization module with ICGM led to increases of 0.8 in \( AP \) and 0.2 in \( AR \). This suggests that using sub-region center points feature for generating the initial outline simplifies subsequent contour optimization, resulting in contours that more accurately represent building outlines. The inclusion of COM further improved the final contours significantly. Compared to the baseline, \( AP \) increased by 2, \( AR \) by 0.2, and \( AP_{75} \) by 1.6. This demonstrates that COM substantially enhances contour quality by capturing dependencies between different vertices, extracting both global and local features, and generating more accurate offsets, thereby producing more precise building outlines. 

\subsection{Discussion}
\begin{figure}[]
    \centering
    \includegraphics[width=9cm]{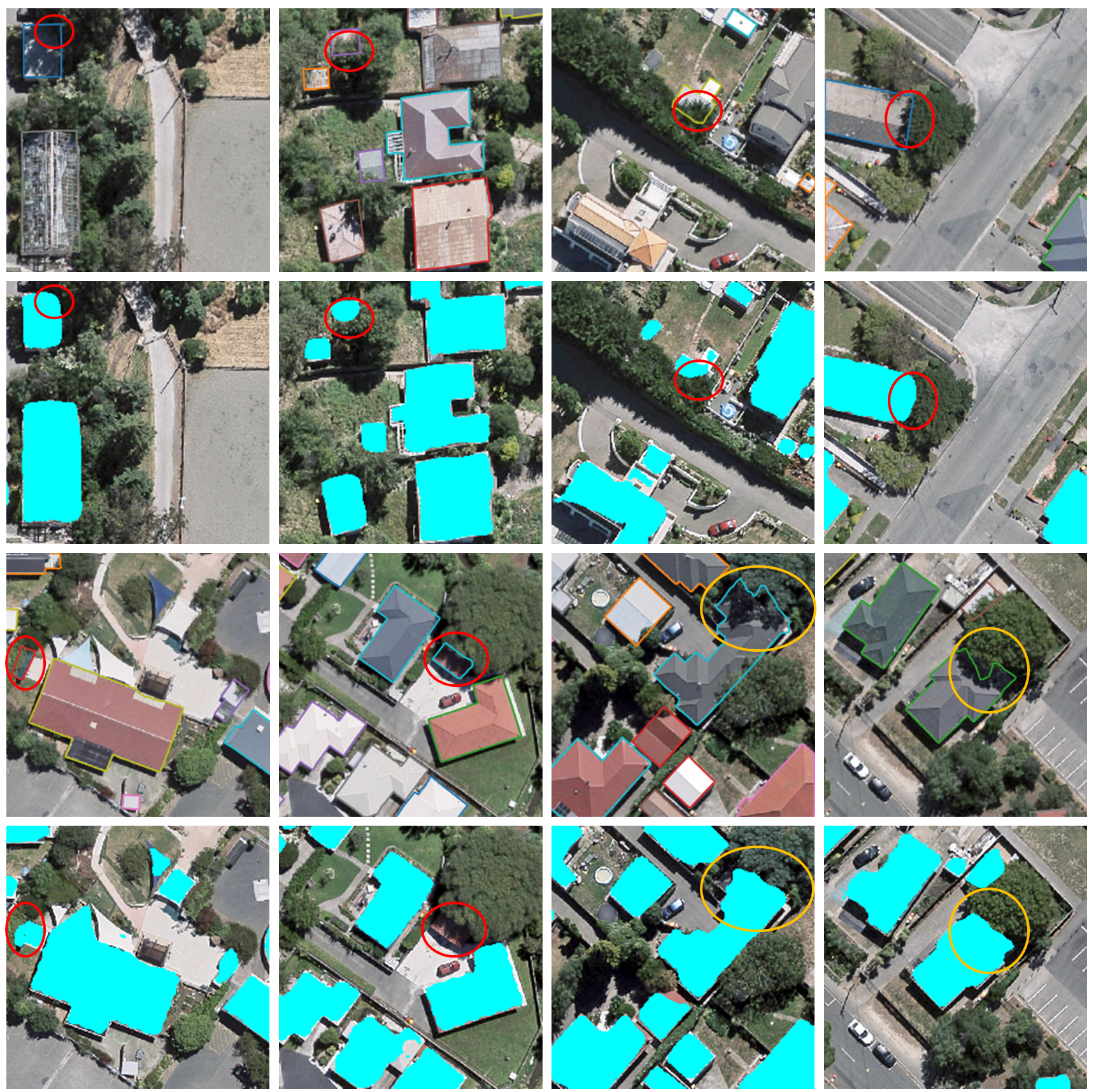}
    \vspace{-2em} % 根据需要调整这个值
    \caption{Examples of prediction results of PolyBuild and YOLACT under occlusion. The occluded region of the building is highlighted by the red circle. When the occluded region is a complex topological structure, the occluded region is marked as a yellow circle. The first two rows are minor occlusion examples, and the last two rows are major occlusion examples.}
    \label{fig:zhedang}
\end{figure}

Classic mask-based method YOLACT struggles to fully extract building contours under occlusion, and the contours are blurred and smooth. PolyBuild exhibits superior performance compared to YOLACT when handling both minor and major occlusions, as demonstrated in Fig. \ref{fig:zhedang}.

For minor occlusions, YOLACT still faces challenges, producing overly smoothed contours or failing to identify the occluded regions, as shown in the first two rows of Fig.~\ref{fig:zhedang}. These issues arise from their reliance on pixel correlations. When pixels are occluded, it is difficult for mask-based methods to discern the true category of the occluded samples. PolyBuild uses sub-region center point features to guide vertex prediction, reducing reliance on a large number of non-occluded pixels. Therefore, PolyBuild can more completely restore building contours in the face of minor occlusions.

For major occlusions, YOLACT often fails to recover complete building structures, resulting in irregular contours or even missed detections. This limitation stems from their inability to infer occluded areas without clear visual feature. In contrast, PolyBuild exploits the geometric relationships between contour vertices and iteratively optimizes their positions based on both local features and global structure, thereby maximizing topological integrity even when large regions of a building are occluded. While PolyBuild demonstrates the capability to reconstruct building contours under occlusion, it struggles to accurately restore the contour of the occluded region when its topology is complex, as shown in the last two columns of the third row of Fig.~\ref{fig:zhedang}. In such cases, the occluded region contains multiple contour corners, and with limited vertex features available, it becomes challenging to precisely localize these occluded corner vertices, which leads to increased instability in the reconstructed contour.

Therefore, future research could explore integrating RNNs and Transformers to more effectively use the known vertex information. Transformers can extract global contour vertex features, while RNNs can enhance the model’s ability to infer the position of subsequent vertices based on the known vertex locations. Global features can assist the RNN in using the spatial relationships among visible vertices to deduce the structure of occluded regions, thereby enabling a more complete and precise restoration of the overall geometric shape of buildings and enhancing the model’s robustness.

\begin{table}[ht]
\centering
{\setlength{\tabcolsep}{2mm}
\caption{The accuracy of building contour extraction at different stages.}
\label{5}
\begin{tabular}{cccccc}
\toprule
Dataset &Contour & \( AP \) (\%) & \( AP_{50} \) (\%) & \( AP_{75} \) (\%) & \( AR \) (\%)  \\ 
\midrule
\multirow{3}{*}{\textit{WHU}}       & \textit{initial}          & 66.9            & 88.8                 & 77.0                 & 72.1              \\ 
                 & \textit{coarse}           & 70.1            & 89.6                 & 79.1                 & 74.7              \\ 
                 & \textit{final}         & 74.0            & 91.1                 & 82.5                 & 80.7              \\ 
\midrule
\multirow{3}{*}{\textit{MIX I}}        & \textit{initial}          & 52.0            & 82.4                 & 58.8                 & 58.3              \\ 
                 & \textit{coarse }          & 54.9            & 83.1                 & 61.6                 & 61.5              \\ 
                 & \textit{final}          & 57.6            & 83.3                 & 64.7                 & 65.2              \\ 
\midrule
\multirow{3}{*}{\textit{MIX II}}       & \textit{initial}          & 45.3            & 76.0                 & 48.5                 & 54.1              \\ 
                 & \textit{coarse}           & 48.1            & 76.7                 & 51.6                 & 57.5              \\ 
                 & \textit{final}          & 49.4            & 76.6                 & 53.1                 & 59.1              \\ 
\bottomrule
\end{tabular}}
\label{table:contour_metrics}
\end{table}

PolyBuild employs a multi-stage optimization approach for extracting building polygons. Table \ref{5} delineates the extraction accuracy at various stages across three datasets. PolyBuild continuously optimizes the contour quality, and the predicted initial contour accuracy can even reach the highest accuracy of some methods.

% As depicted in fig.10, the process initiates with generating an initial contour that approximates the building's shape. Through subsequent network optimizations, this contour is refined to tightly conform to the building instance, progressively enhancing contour accuracy and ultimately yielding building polygon contours that rival those produced by manual annotation .

\begin{figure*}[]
    \centering
    \includegraphics[width=18cm]{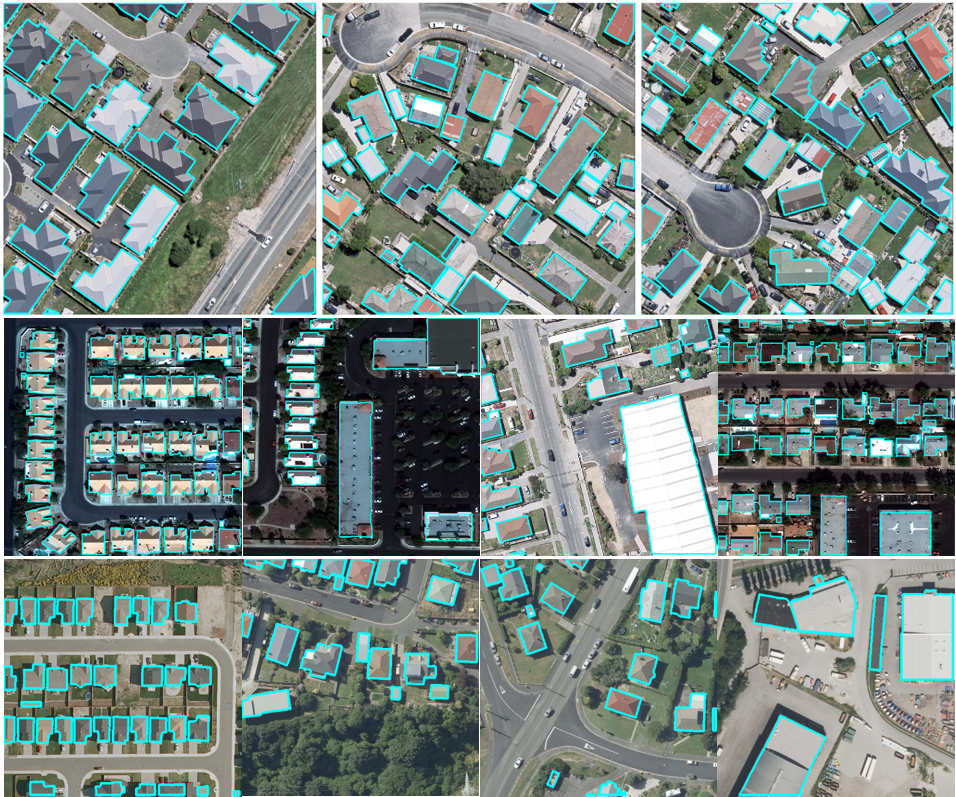}
    \caption{Qualitative results obtained by PolyBuild. From top to bottom: results on the WHU aerial building dataset, WHU-Mix test set I, WHU-Mix test set II.}
    \label{10}
\end{figure*}

Given a remote sensing image, PolyBuild can generate building contour polygons end-to-end without additional post-processing operations. Fig. \ref{10} shows the extraction results of PolyBuild on the WHU aerial building dataset, WHU-Mix test set I, and WHU-Mix test set II. As shown in the figure, PolyBuild accurately captures building features under three different viewing conditions, thereby generating extremely high-quality building outline polygons.

Although PolyBuild integrates CNN-Transformer, its parameter is only 30.3M. We further test the inference speed of PolyBuild, and under the resource configuration of an RTX 2080 Ti GPU, it achieved an inference speed of 30.46 FPS on 512×512 images.

\subsection{Limitations and future work}
\begin{figure}[]
    \centering
    % \vspace{-1em} % 根据需要调整这个值
    \includegraphics[width=9cm]{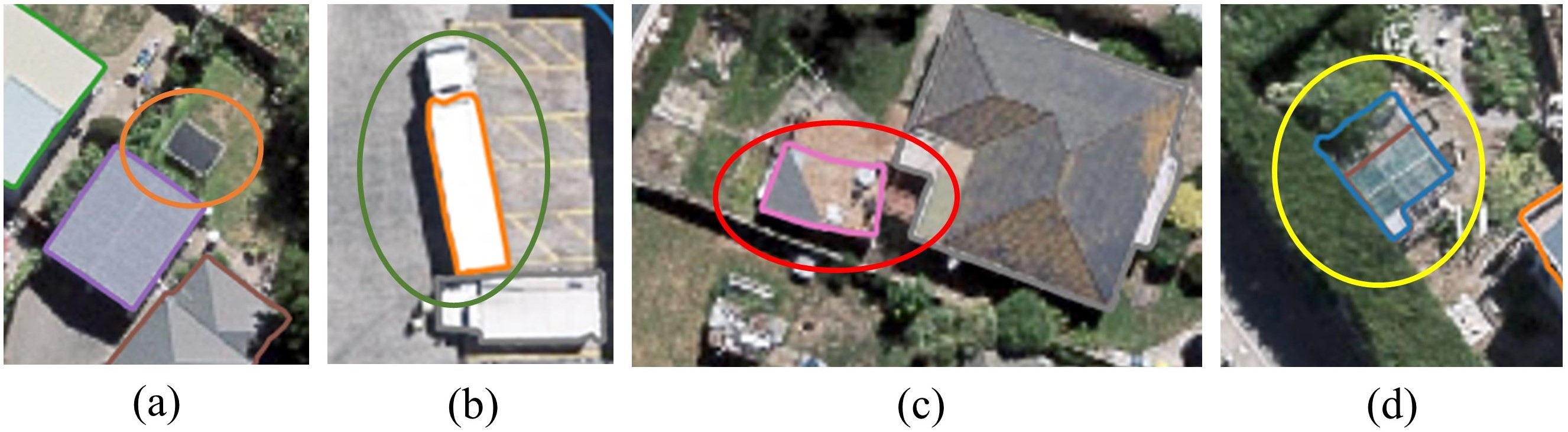}
    \caption{Examples of buildings that PolyBuild fails to extract correctly. (a) Missed detection of a building; (b) A truck misidentified as a building; (c) Abandoned land misidentified as a building; (d) Duplicate detection of a building.}
    \label{limition}
\end{figure}

Like most contour-based methods, PolyBuild first performs object detection and then applies contour extraction based on the detected instances. However, this two-stage approach has a drawback: the accuracy of contour extraction is dependent on the precision of object detection. PolyBuild uses the CenterNet detector to locate building targets, and due to the limitations of the detector's performance, PolyBuild experiences a certain degree of missed detections and false positives. For example, in Fig. \ref{limition}(a), small buildings are not correctly identified; in Fig. \ref{limition}(b), a large truck is mistakenly detected as a building; in Fig. \ref{limition}(c), abandoned land is incorrectly classified as a building; and in Fig. \ref{limition}(d), the same building is detected multiple times.

In future work, we plan to apply more powerful detectors to contour extraction to further improve the performance.

\section{Conclusion}
This study proposes PolyBuild, an advanced and elegantly structured end-to-end building contour extraction method. PolyBuild introduces a novel contour generation strategy to address the issue of contour uncertainty when regressing from a single center point feature. PolyBuild regresses the initial contour based on the coupled sub-region center point features, providing rich contour information and thus reducing the difficulty of network learning. Additionally, to automatically capture dependencies among contour points, we designed a contour optimization module that integrates CNN and Transformer architectures. This module leverages both CNN and Transformer capabilities to extract global and local features of contours, capture complex semantic relationships between contour vertices, and achieve precise building contour extraction. PolyBuild demonstrates state-of-the-art (SOTA) performance on the WHU aerial dataset and the WHU-Mix dataset, providing high-precision building contour polygons at the instance level to meet the needs of the downstream surveying and mapping industry.

\bibliographystyle{IEEEtran}
\small\bibliography{referce}

@inproceedings{zorzi2022polyworld,
  title={Polyworld: Polygonal building extraction with graph neural networks in satellite images},
  author={Zorzi, Stefano and Bazrafkan, Shabab and Habenschuss, Stefan and Fraundorfer, Friedrich},
  booktitle={Proceedings of the IEEE/CVF Conference on Computer Vision and Pattern Recognition},
  pages={1848--1857},
  year={2022}
}

@article{xu2021gated,
  title={Gated spatial memory and centroid-aware network for building instance extraction},
  author={Xu, Lele and Li, Ye and Xu, Jinzhong and Guo, Lili},
  journal={IEEE Transactions on Geoscience and Remote Sensing},
  volume={60},
  pages={1--14},
  year={2021},
  publisher={IEEE}
}

@article{li2014extracting,
  title={Extracting man-made objects from high spatial resolution remote sensing images via fast level set evolutions},
  author={Li, Zhongbin and Shi, Wenzhong and Wang, Qunming and Miao, Zelang},
  journal={IEEE Transactions on Geoscience and Remote Sensing},
  volume={53},
  number={2},
  pages={883--899},
  year={2014},
  publisher={IEEE}
}

@inproceedings{carion2020end,
  title={End-to-end object detection with transformers},
  author={Carion, Nicolas and Massa, Francisco and Synnaeve, Gabriel and Usunier, Nicolas and Kirillov, Alexander and Zagoruyko, Sergey},
  booktitle={European conference on computer vision},
  pages={213--229},
  year={2020},
  organization={Springer}
}

@article{wei2024lines,
  title={From lines to Polygons: Polygonal building contour extraction from High-Resolution remote sensing imagery},
  author={Wei, Shiqing and Zhang, Tao and Yu, Dawen and Ji, Shunping and Zhang, Yongjun and Gong, Jianya},
  journal={ISPRS Journal of Photogrammetry and Remote Sensing},
  volume={209},
  pages={213--232},
  year={2024},
  publisher={Elsevier}
}

@article{wang2024sampolybuild,
  title={SAMPolyBuild: Adapting the Segment Anything Model for polygonal building extraction},
  author={Wang, Chenhao and Chen, Jingbo and Meng, Yu and Deng, Yupeng and Li, Kai and Kong, Yunlong},
  journal={ISPRS Journal of Photogrammetry and Remote Sensing},
  volume={218},
  pages={707--720},
  year={2024},
  publisher={Elsevier}
}

@article{li2023joint,
  title={Joint semantic--geometric learning for polygonal building segmentation from high-resolution remote sensing images},
  author={Li, Weijia and Zhao, Wenqian and Yu, Jinhua and Zheng, Juepeng and He, Conghui and Fu, Haohuan and Lin, Dahua},
  journal={ISPRS Journal of Photogrammetry and Remote Sensing},
  volume={201},
  pages={26--37},
  year={2023},
  publisher={Elsevier}
}

@article{xu2023hisup,
  title={HiSup: Accurate polygonal mapping of buildings in satellite imagery with hierarchical supervision},
  author={Xu, Bowen and Xu, Jiakun and Xue, Nan and Xia, Gui-Song},
  journal={ISPRS Journal of Photogrammetry and Remote Sensing},
  volume={198},
  pages={284--296},
  year={2023},
  publisher={Elsevier}
}

@article{guo2020scene,
  title={Scene-driven multitask parallel attention network for building extraction in high-resolution remote sensing images},
  author={Guo, Haonan and Shi, Qian and Du, Bo and Zhang, Liangpei and Wang, Dongzhi and Ding, Huaxiang},
  journal={IEEE Transactions on Geoscience and Remote Sensing},
  volume={59},
  number={5},
  pages={4287--4306},
  year={2020},
  publisher={IEEE}
}

@book{hart2000pattern,
  title={Pattern classification},
  author={Hart, Peter E and Stork, David G and Duda, Richard O and others},
  year={2000},
  publisher={Wiley Hoboken}
}

@article{shrivastava2015automatic,
  title={Automatic building extraction based on multiresolution segmentation using remote sensing data},
  author={Shrivastava, Neeti and Kumar Rai, Praveen},
  journal={Geographia Polonica},
  volume={88},
  number={3},
  pages={407--421},
  year={2015},
  publisher={IGiPZ PAN}
}

@article{canny1986computational,
  title={A computational approach to edge detection},
  author={Canny, John},
  journal={IEEE Transactions on pattern analysis and machine intelligence},
  number={6},
  pages={679--698},
  year={1986},
  publisher={Ieee}
}

@article{yu2021automatic,
  title={Automatic 3D building reconstruction from multi-view aerial images with deep learning},
  author={Yu, Dawen and Ji, Shunping and Liu, Jin and Wei, Shiqing},
  journal={ISPRS Journal of Photogrammetry and Remote Sensing},
  volume={171},
  pages={155--170},
  year={2021},
  publisher={Elsevier}
}

@article{guo2023decoupling,
  title={Decoupling semantic and edge representations for building footprint extraction from remote sensing images},
  author={Guo, Haonan and Su, Xin and Wu, Chen and Du, Bo and Zhang, Liangpei},
  journal={IEEE Transactions on Geoscience and Remote Sensing},
  volume={61},
  pages={1--16},
  year={2023},
  publisher={IEEE}
}

@article{wei2019toward,
  title={Toward automatic building footprint delineation from aerial images using CNN and regularization},
  author={Wei, Shiqing and Ji, Shunping and Lu, Meng},
  journal={IEEE Transactions on Geoscience and Remote Sensing},
  volume={58},
  number={3},
  pages={2178--2189},
  year={2019},
  publisher={IEEE}
}

@article{zhou2022bomsc,
  title={BOMSC-Net: Boundary optimization and multi-scale context awareness based building extraction from high-resolution remote sensing imagery},
  author={Zhou, Yuan and Chen, Zhanlong and Wang, Bin and Li, Shuangjiang and Liu, Hao and Xu, Daozhu and Ma, Chao},
  journal={IEEE Transactions on Geoscience and Remote Sensing},
  volume={60},
  pages={1--17},
  year={2022},
  publisher={IEEE}
}

@article{zhu2020map,
  title={MAP-Net: Multiple attending path neural network for building footprint extraction from remote sensed imagery},
  author={Zhu, Qing and Liao, Cheng and Hu, Han and Mei, Xiaoming and Li, Haifeng},
  journal={IEEE Transactions on Geoscience and Remote Sensing},
  volume={59},
  number={7},
  pages={6169--6181},
  year={2020},
  publisher={IEEE}
}

@article{hu2023boundary,
  title={Boundary shape-preserving model for building mapping from high-resolution remote sensing images},
  author={Hu, Anna and Wu, Liang and Chen, Siqiong and Xu, Yongyang and Wang, Haitao and Xie, Zhong},
  journal={IEEE Transactions on Geoscience and Remote Sensing},
  volume={61},
  pages={1--17},
  year={2023},
  publisher={IEEE}
}

@inproceedings{he2017mask,
  title={Mask r-cnn},
  author={He, Kaiming and Gkioxari, Georgia and Doll{\'a}r, Piotr and Girshick, Ross},
  booktitle={Proceedings of the IEEE international conference on computer vision},
  pages={2961--2969},
  year={2017}
}

@inproceedings{bolya2019yolact,
  title={Yolact: Real-time instance segmentation},
  author={Bolya, Daniel and Zhou, Chong and Xiao, Fanyi and Lee, Yong Jae},
  booktitle={Proceedings of the IEEE/CVF international conference on computer vision},
  pages={9157--9166},
  year={2019}
}

@inproceedings{xie2020polarmask,
  title={Polarmask: Single shot instance segmentation with polar representation},
  author={Xie, Enze and Sun, Peize and Song, Xiaoge and Wang, Wenhai and Liu, Xuebo and Liang, Ding and Shen, Chunhua and Luo, Ping},
  booktitle={Proceedings of the IEEE/CVF conference on computer vision and pattern recognition},
  pages={12193--12202},
  year={2020}
}

@inproceedings{peng2020deep,
  title={Deep snake for real-time instance segmentation},
  author={Peng, Sida and Jiang, Wen and Pi, Huaijin and Li, Xiuli and Bao, Hujun and Zhou, Xiaowei},
  booktitle={Proceedings of the IEEE/CVF conference on computer vision and pattern recognition},
  pages={8533--8542},
  year={2020}
}

@inproceedings{liu2021dance,
  title={Dance: A deep attentive contour model for efficient instance segmentation},
  author={Liu, Zichen and Liew, Jun Hao and Chen, Xiangyu and Feng, Jiashi},
  booktitle={Proceedings of the IEEE/CVF winter conference on applications of computer vision},
  pages={345--354},
  year={2021}
}

@inproceedings{zhang2022e2ec,
  title={E2ec: An end-to-end contour-based method for high-quality high-speed instance segmentation},
  author={Zhang, Tao and Wei, Shiqing and Ji, Shunping},
  booktitle={Proceedings of the IEEE/CVF conference on computer vision and pattern recognition},
  pages={4443--4452},
  year={2022}
}

@article{wei2023buildmapper,
  title={BuildMapper: A fully learnable framework for vectorized building contour extraction},
  author={Wei, Shiqing and Zhang, Tao and Ji, Shunping and Luo, Muying and Gong, Jianya},
  journal={ISPRS Journal of Photogrammetry and Remote Sensing},
  volume={197},
  pages={87--104},
  year={2023},
  publisher={Elsevier}
}

@inproceedings{ling2019fast,
  title={Fast interactive object annotation with curve-gcn},
  author={Ling, Huan and Gao, Jun and Kar, Amlan and Chen, Wenzheng and Fidler, Sanja},
  booktitle={Proceedings of the IEEE/CVF conference on computer vision and pattern recognition},
  pages={5257--5266},
  year={2019}
}

@article{wei2021concentric,
  title={A concentric loop convolutional neural network for manual delineation-level building boundary segmentation from remote-sensing images},
  author={Wei, Shiqing and Zhang, Tao and Ji, Shunping},
  journal={IEEE Transactions on Geoscience and Remote Sensing},
  volume={60},
  pages={1--11},
  year={2021},
  publisher={IEEE}
}

@inproceedings{ronneberger2015u,
  title={U-net: Convolutional networks for biomedical image segmentation},
  author={Ronneberger, Olaf and Fischer, Philipp and Brox, Thomas},
  booktitle={Medical image computing and computer-assisted intervention--MICCAI 2015: 18th international conference, Munich, Germany, October 5-9, 2015, proceedings, part III 18},
  pages={234--241},
  year={2015},
  organization={Springer}
}

@inproceedings{sun2019deep,
  title={Deep high-resolution representation learning for human pose estimation},
  author={Sun, Ke and Xiao, Bin and Liu, Dong and Wang, Jingdong},
  booktitle={Proceedings of the IEEE/CVF conference on computer vision and pattern recognition},
  pages={5693--5703},
  year={2019}
}

@article{jung2021boundary,
  title={Boundary enhancement semantic segmentation for building extraction from remote sensed image},
  author={Jung, Hoin and Choi, Han-Soo and Kang, Myungjoo},
  journal={IEEE Transactions on Geoscience and Remote Sensing},
  volume={60},
  pages={1--12},
  year={2021},
  publisher={IEEE}
}

@article{xu2023bctnet,
  title={BCTNet: Bi-branch cross-fusion transformer for building footprint extraction},
  author={Xu, Lele and Li, Ye and Xu, Jinzhong and Zhang, Yue and Guo, Lili},
  journal={IEEE Transactions on Geoscience and Remote Sensing},
  volume={61},
  pages={1--14},
  year={2023},
  publisher={IEEE}
}

@article{dosovitskiy2020image,
  title={An image is worth 16x16 words: Transformers for image recognition at scale},
  author={Dosovitskiy, Alexey and Beyer, Lucas and Kolesnikov, Alexander and Weissenborn, Dirk and Zhai, Xiaohua and Unterthiner, Thomas and Dehghani, Mostafa and Minderer, Matthias and Heigold, Georg and Gelly, Sylvain and others},
  journal={arXiv preprint arXiv:2010.11929},
  year={2020}
}

@article{xie2021segformer,
  title={SegFormer: Simple and efficient design for semantic segmentation with transformers},
  author={Xie, Enze and Wang, Wenhai and Yu, Zhiding and Anandkumar, Anima and Alvarez, Jose M and Luo, Ping},
  journal={Advances in neural information processing systems},
  volume={34},
  pages={12077--12090},
  year={2021}
}

@inproceedings{liu2018path,
  title={Path aggregation network for instance segmentation},
  author={Liu, Shu and Qi, Lu and Qin, Haifang and Shi, Jianping and Jia, Jiaya},
  booktitle={Proceedings of the IEEE conference on computer vision and pattern recognition},
  pages={8759--8768},
  year={2018}
}

@inproceedings{tian2020conditional,
  title={Conditional convolutions for instance segmentation},
  author={Tian, Zhi and Shen, Chunhua and Chen, Hao},
  booktitle={Computer Vision--ECCV 2020: 16th European Conference, Glasgow, UK, August 23--28, 2020, Proceedings, Part I 16},
  pages={282--298},
  year={2020},
  organization={Springer}
}

@inproceedings{chen2020blendmask,
  title={Blendmask: Top-down meets bottom-up for instance segmentation},
  author={Chen, Hao and Sun, Kunyang and Tian, Zhi and Shen, Chunhua and Huang, Yongming and Yan, Youliang},
  booktitle={Proceedings of the IEEE/CVF conference on computer vision and pattern recognition},
  pages={8573--8581},
  year={2020}
}

@article{huang2021sequentially,
  title={Sequentially delineation of rooftops with holes from VHR aerial images using a convolutional recurrent neural network},
  author={Huang, Wei and Liu, Zeping and Tang, Hong and Ge, Jiayi},
  journal={Remote Sensing},
  volume={13},
  number={21},
  pages={4271},
  year={2021},
  publisher={MDPI}
}

@article{zhao2021building,
  title={Building outline delineation: From aerial images to polygons with an improved end-to-end learning framework},
  author={Zhao, Wufan and Persello, Claudio and Stein, Alfred},
  journal={ISPRS journal of photogrammetry and remote sensing},
  volume={175},
  pages={119--131},
  year={2021},
  publisher={Elsevier}
}

@article{liu2022building,
  title={Building outline delineation from VHR remote sensing images using the convolutional recurrent neural network embedded with line segment information},
  author={Liu, Zeping and Tang, Hong and Huang, Wei},
  journal={IEEE Transactions on Geoscience and Remote Sensing},
  volume={60},
  pages={1--13},
  year={2022},
  publisher={IEEE}
}

@article{zhou2019objects,
  title={Objects as points},
  author={Zhou, Xingyi and Wang, Dequan and Kr{\"a}henb{\"u}hl, Philipp},
  journal={arXiv preprint arXiv:1904.07850},
  year={2019}
}

@article{ji2018fully,
  title={Fully convolutional networks for multisource building extraction from an open aerial and satellite imagery data set},
  author={Ji, Shunping and Wei, Shiqing and Lu, Meng},
  journal={IEEE Transactions on geoscience and remote sensing},
  volume={57},
  number={1},
  pages={574--586},
  year={2018},
  publisher={IEEE}
}

@misc{yap2018open,
  title={open ai tanzania building footprint segmentation challenge,”},
  author={Yap, J},
  year={2018}
}

@misc{mohanty2018crowdai,
  title={Crowdai dataset},
  author={Mohanty, Sharada Prasanna},
  year={2018}
}

@article{van2018spacenet,
  title={Spacenet: A remote sensing dataset and challenge series},
  author={Van Etten, Adam and Lindenbaum, Dave and Bacastow, Todd M},
  journal={arXiv preprint arXiv:1807.01232},
  year={2018}
}

@inproceedings{maggiori2017can,
  title={Can semantic labeling methods generalize to any city? the inria aerial image labeling benchmark},
  author={Maggiori, Emmanuel and Tarabalka, Yuliya and Charpiat, Guillaume and Alliez, Pierre},
  booktitle={2017 IEEE International geoscience and remote sensing symposium (IGARSS)},
  pages={3226--3229},
  year={2017},
  organization={IEEE}
}

@inproceedings{wang2020solo,
  title={Solo: Segmenting objects by locations},
  author={Wang, Xinlong and Kong, Tao and Shen, Chunhua and Jiang, Yuning and Li, Lei},
  booktitle={Computer Vision--ECCV 2020: 16th European Conference, Glasgow, UK, August 23--28, 2020, Proceedings, Part XVIII 16},
  pages={649--665},
  year={2020},
  organization={Springer}
}

@inproceedings{cheng2022masked,
  title={Masked-attention mask transformer for universal image segmentation},
  author={Cheng, Bowen and Misra, Ishan and Schwing, Alexander G and Kirillov, Alexander and Girdhar, Rohit},
  booktitle={Proceedings of the IEEE/CVF conference on computer vision and pattern recognition},
  pages={1290--1299},
  year={2022}
}

@misc{lyu2022rtmdet,
      title={RTMDet: An Empirical Study of Designing Real-Time Object Detectors},
      author={Chengqi Lyu and Wenwei Zhang and Haian Huang and Yue Zhou and Yudong Wang and Yanyi Liu and Shilong Zhang and Kai Chen},
      year={2022},
      eprint={2212.07784},
      archivePrefix={arXiv},
      primaryClass={cs.CV}
}

@article{yunfeng2018extraction,
  title={The Extraction of Building Shadow and the Estimation of Building Heights Based on Morphology and Spectral Characteristic Parameters},
  author={Yunfeng, HU and Qianli, ZHANG},
  journal={Bulletin of Surveying and Mapping},
  number={6},
  pages={22},
  year={2018}
}

@article{huang2011multidirectional,
  title={A multidirectional and multiscale morphological index for automatic building extraction from multispectral GeoEye-1 imagery},
  author={Huang, Xin and Zhang, Liangpei},
  journal={Photogrammetric Engineering \& Remote Sensing},
  volume={77},
  number={7},
  pages={721--732},
  year={2011},
  publisher={American Society for Photogrammetry and Remote Sensing}
}

@inproceedings{lin2017focal,
  title={Focal loss for dense object detection},
  author={Lin, Tsung-Yi and Goyal, Priya and Girshick, Ross and He, Kaiming and Doll{\'a}r, Piotr},
  booktitle={Proceedings of the IEEE international conference on computer vision},
  pages={2980--2988},
  year={2017}
}

@article{yasir2024shipgeonet,
  title={ShipGeoNet: SAR image-based geometric feature extraction of ships using convolutional neural networks},
  author={Yasir, Muhammad and Liu, Shanwei and Mingming, Xu and Wan, Jianhua and Pirasteh, Saied and Dang, Kinh Bac},
  journal={IEEE Transactions on Geoscience and Remote Sensing},
  year={2024},
  publisher={IEEE}
}

@inproceedings{zhao2020building,
  title={Building instance segmentation and boundary regularization from high-resolution remote sensing images},
  author={Zhao, Wufan and Persello, Claudio and Stein, Alfred},
  booktitle={IGARSS 2020-2020 IEEE International Geoscience and Remote Sensing Symposium},
  pages={3916--3919},
  year={2020},
  organization={IEEE}
}

@inproceedings{li2019topological,
  title={Topological map extraction from overhead images},
  author={Li, Zuoyue and Wegner, Jan Dirk and Lucchi, Aur{\'e}lien},
  booktitle={Proceedings of the IEEE/CVF International Conference on Computer Vision},
  pages={1715--1724},
  year={2019}
}

@ARTICLE{10824925,
  author={Zhao, Zhe and Zhao, Boya and Wu, Yuanfeng and He, Zutian and Gao, Lianru},
  journal={IEEE Journal of Selected Topics in Applied Earth Observations and Remote Sensing}, 
  title={Building extraction from high-resolution multispectral and SAR images using a boundary-link multimodal fusion network}, 
  year={2025},
  volume={},
  number={},
  pages={1-15},
  doi={10.1109/JSTARS.2025.3525709}}

\end{document}